\documentclass[11pt]{article}

\usepackage{acl}
\usepackage{times}
\usepackage{latexsym}
\usepackage[T1]{fontenc}
\usepackage[utf8]{inputenc}
\usepackage{microtype}
\usepackage{inconsolata}
\usepackage{graphicx}
\usepackage{placeins}
\usepackage{booktabs}
\usepackage{array}
\usepackage{colortbl}
\usepackage{amsmath,amssymb}
\usepackage{algorithm}
\usepackage{algpseudocode}
\usepackage{multirow}
\usepackage{amsmath,amsfonts,bm}

\def\eqref#1{equation~\ref{#1}}

\def\1{\bm{1}}

\DeclareMathAlphabet{\mathsfit}{\encodingdefault}{\sfdefault}{m}{sl}
\SetMathAlphabet{\mathsfit}{bold}{\encodingdefault}{\sfdefault}{bx}{n}

\title{Batch-Paced: Fixed-Budget Batch Decoding\\
in Discrete Diffusion Language Models}

\author{%
Shuai Zhang\textsuperscript{1,2} \quad
Zhanchao Zhou\textsuperscript{2} \quad
Huachuan Qiu\textsuperscript{2} \quad
Anqi Li\textsuperscript{2} \quad
Zhenzhong Lan\textsuperscript{2}\thanks{Corresponding author.}\\
\textsuperscript{1}Zhejiang University \qquad
\textsuperscript{2}Westlake University%
}

\newcommand{\ours}{Batch-Paced}

\newenvironment{compactequation}{%
  \setlength{\abovedisplayskip}{5pt}%
  \setlength{\belowdisplayskip}{5pt}%
  \setlength{\abovedisplayshortskip}{5pt}%
  \setlength{\belowdisplayshortskip}{5pt}%
  \begin{equation}%
}{\end{equation}}

\begin{document}
\maketitle

\begin{abstract}
As generative language models become more capable and widely used, inference demand grows, making
efficient serving increasingly important. Discrete diffusion language models (dLLMs) offer a promising
path to more efficient inference through parallel token generation. Prior dLLM decoding methods,
however, focus on improving inference efficiency for individual sequences. We formulate fixed-budget
dLLM batch decoding and introduce \ours{}, a training-free method that jointly optimizes the allocation
of decoding compute across requests to improve overall batch quality.
Across two dLLM backbones, five benchmarks and their representative mixtures, \ours{} improves batch
performance under matched budgets; gains generally grow with batch size and largely persist
across decoding configurations. Analysis of requests' decoding trajectories suggests that these gains
arise largely by raising the committed-token confidence of lower-confidence requests.
We further show that \ours{}'s batch decode time remains competitive and that adding a decoding-depth
allowance during Batch-Paced decoding can shorten it further at a modest performance cost.
\end{abstract}

\section{Introduction}

Generative language models now support applications ranging from conversational assistants to code
generation \citep{yu2022orca}. A Microsoft 365 workload study reports more than 10 million LLM requests per
day, underscoring the need for efficient inference serving \citep{jaiswal2025sageserve}. Commercial APIs reflect
this need through explicit rate limits: OpenAI's vary by model and apply at the organization
or project level, while Anthropic's specify request and token limits by model and usage tier
\citep{openaiscale,anthropiclimits}. In actual deployments, the systems behind these APIs must
schedule concurrent requests over finite GPU capacity. Sarathi-Serve, for example, limits the number
of tokens processed per iteration to balance latency and throughput \citep{agrawal2024taming}. This motivates a
batch-decoding question: given a fixed decoding-compute allowance for a batch, how should it be
allocated across requests to maximize the batch's overall output quality?

Diffusion language models (dLLMs) \citep{nie2026large,ye2508dream,hu2026elf,zhang2026low} offer a promising route to more
efficient generation by predicting multiple masked positions in parallel. This parallel-generation
efficiency has motivated diffusion-based speculative decoding
\citep{chen2026dflash,cheng2026dspark,zhang2026cost}. Subsequent work on dLLM decoding uses signals such as token confidence,
entropy, temporal consistency, and denoising history to decide when to commit tokens for a single
sequence, aiming to improve individual-sequence decoding efficiency
\citep{wu2026fast,ben2026accelerated,kim2026klass,wang2026creditdecoding}. However, fixed-budget dLLM batch decoding is underexplored.

\begin{figure*}[!t]
\centering
\includegraphics[width=0.99\textwidth]{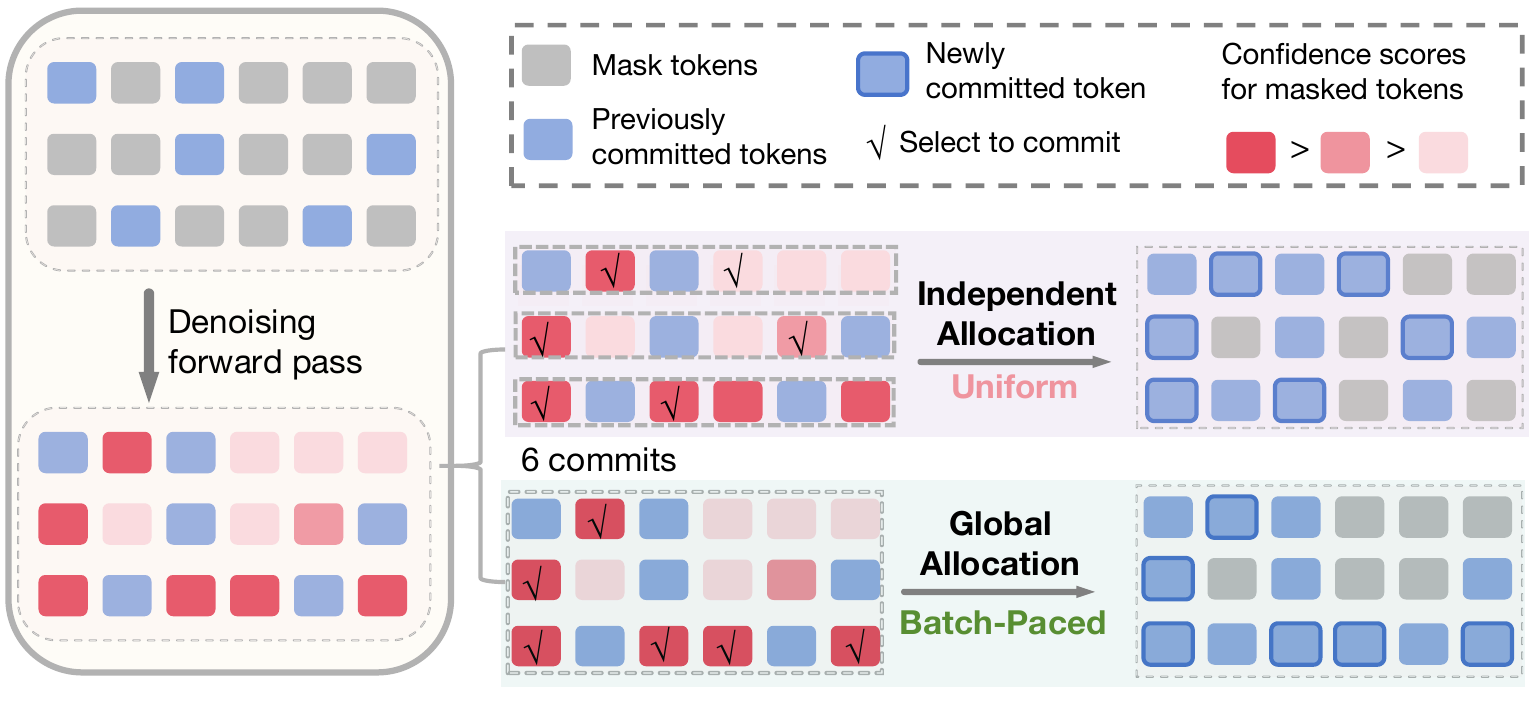}
\caption{\textbf{Batch-Paced uses global compute allocation within the batch rather than independent sequence-wise allocation.}
After a denoising forward pass assigns confidence scores to masked-token predictions, Uniform commits
a fixed number of positions per sequence, whereas Batch-Paced ranks positions across the batch.}
\label{fig:method-overview}
\end{figure*}

We introduce \ours{}, a training-free method for fixed-budget dLLM batch decoding. It jointly ranks
masked-token confidence across requests to distribute commits, allowing requests to advance at
different paces while meeting the batch's compute budget. Figure~\ref{fig:method-overview}
illustrates this allocation contrast.
We evaluate \ours{} on five benchmarks
and representative mixtures using two discrete diffusion language models. At matched denoising
budgets, it improves over Uniform, which uses independent sequence-wise allocation and commits the
same fixed number of tokens for every request at each denoising pass; gains generally grow with batch
size. Our analysis suggests that this batch-level improvement comes primarily from raising the confidence trajectory of
lower-confidence requests. To see whether this performance pattern depends on a particular decoding
setup, we vary block length, generation length, and confidence-score function; the gains largely persist across
these configurations. We also compare Batch-Paced with Fast-dLLM, a strong adaptive baseline. Their
performance is broadly comparable, although matching a target batch budget with Fast-dLLM requires
additional calibration runs. Since batch decode time also matters, we measure it alongside
performance: Batch-Paced is broadly comparable to Uniform but
generally slower than Fast-dLLM. We extend Batch-Paced with a decoding-depth allowance that limits the
number of batched denoising steps, bringing batch decode time closer to Fast-dLLM's at some performance
cost. Our contributions are:
\begin{itemize}
\item We pose fixed-budget dLLM batch decoding as an underexplored problem and propose \ours{} to
jointly allocate compute across requests for improved batch quality.
\item Empirically, \ours{} improves batch performance under fixed denoising budgets largely by raising
the committed-token confidence of lower-confidence requests.
\item We show that \ours{}'s gains largely persist across decoding configurations and that its batch
decode time remains competitive; a \ours{} extension can further shorten batch decoding at a modest
performance cost.
\end{itemize}

\section{Related Work}

\subsection{Decoding Policies for Discrete Diffusion Language Models}

Discrete diffusion language models generate through iterative parallel denoising \citep{nie2026large,ye2508dream}.
Block diffusion adopts a semi-autoregressive structure, with diffusion within each
block and autoregressive generation across blocks \citep{arriola2025block,bie2025llada2,team2026diffusiongemma}. Recent work improves dLLM
inference by adapting token commitments within a sequence. These methods draw on different signals:
Fast-dLLM uses token confidence, EB-Sampler bounds
unmasking entropy, KLASS uses temporal KL, and CreditDecoding accumulates denoising history
\citep{wu2026fast,ben2026accelerated,kim2026klass,wang2026creditdecoding}. Dynamic-dLLM combines adaptive parallel decoding with
layer-wise cache-update budgets \citep{wu2026dynamic}. Other approaches incorporate an auxiliary
autoregressive score, alternate exploratory and accelerated stages, use answer convergence to guide
commits, or learn token-level unmasking policies
\citep{israel2026accelerating,wei2026accelerating,li2026diffusion,bao2026learning,jazbec2025learning,hong2026improving}. These methods adapt decoding within
individual sequences, but do not consider multiple sequences decoded together as a batch.

\subsection{Inference Compute Allocation}

Research on inference-compute allocation spans several granularities. CALM varies model depth at the
token level \citep{schuster2022confident}. Test-time scaling allocates a problem's compute between parallel
samples and sequential revisions \citep{snell2025scaling}. Strategic Scaling formulates allocation across
queries as an online bandit problem \citep{zuo2026strategic}, while Certaindex sets reasoning-token
budgets across queries based on answer stability \citep{fu2024efficiently}. Dynasor schedules multi-stage
reasoning programs \citep{fu2024efficiently}. At the query-set level, Damani et al. provide the closest
language-model precedent to our setting: they allocate an average compute budget across queries
according to predicted marginal reward \citep{damani2025learning}. We study the analogous problem in batch
dLLM decoding, distributing a fixed denoising budget among sequences decoded together at each pass as
completed sequences release budget.

\subsection{Batched dLLM Inference Systems}

Recent systems improve dLLM serving efficiency.
dLLM-Serve addresses memory use and compute scheduling \citep{fan2026dllm}, while DiLaServe coordinates
request-level decoding thresholds with latency objectives and aggregate cluster load \citep{chang2026dilaserve}.
These systems focus on request execution and service-level objectives. Complementing this system-level
work, we study fixed-budget decoding within an already formed batch, allocating denoising compute across
sequences to improve batch output quality and measuring the resulting synchronized batch decode time.

\section{Preliminaries and Problem Setup}
\label{sec:preliminaries}

\subsection{Discrete Diffusion Decoding}
\label{sec:diffusion-decoding}

A discrete diffusion language model initializes each sequence with a fully masked generation region. At each denoising pass, the model predicts all
remaining masked positions in parallel; the decoder commits a subset and leaves the rest masked. For a sequence with
generation positions $\mathcal G=\{1,\ldots,L\}$, let
$\mathbf{x}$ be its current state and $\mathcal M\subseteq\mathcal G$ its remaining masked positions.
Given the current sequence state, a denoising forward pass produces a distribution
$\mathbf p_j:=p_\theta(\cdot\mid\mathbf{x})_j$ over $\mathcal V$ for each remaining masked position
$j\in\mathcal M$. Greedy decoding yields the corresponding prediction $\hat{x}_j$, while the policy
assigns position $j$ a score $s_j=\phi(\mathbf p_j)$. Collecting these scores as
$\mathbf s=(s_j)_{j\in\mathcal M}$, the predicted tokens and selected set are defined by
\begin{compactequation}
 \hat{x}_j=\arg\max_{v\in\mathcal V} [\mathbf p_j]_v,\;
 \mathcal C=\mathcal S(\mathcal M;\mathbf s).
 \label{eq:commit-score}
\end{compactequation}
Committing $\mathcal C$ updates $x_j\leftarrow\hat{x}_j$ for $j\in\mathcal C$ and
$\mathcal M\leftarrow\mathcal M\setminus\mathcal C$. The policy-specific selector $\mathcal S$ may
use top-$k$, thresholding, or pooled ranking.

\subsection{Fixed-Budget Batch Decoding}
\label{sec:batch-decoding}

In dLLM serving, multiple input requests are processed together as a batch.  A batch decoder
therefore considers the overall denoising progress of the batch when making decoding decisions for
each request sequence.  Under a fixed total denoising budget, the question is whether such batch-level
decisions can improve the final quality of the batch.

Formally, let a batch contain $B$ sequences with a total budget $T=\bar{T}B$, where $\bar{T}$
denotes the average number of denoising forward passes allocated to each sequence.
At a denoising pass, let $\mathcal M_i$ denote the remaining masked generation positions of sequence
$i$.  The batch's candidate positions are
\begin{compactequation}
 \mathcal E=\{(i,j):i\in\{1,\ldots,B\},\ j\in\mathcal M_i\}.
 \label{eq:batch-candidates}
\end{compactequation}
For each candidate $(i,j)\in\mathcal E$, the batched forward pass gives a score
$s_{ij}=\phi(\mathbf p_{ij})$.  For sequence $i$, collect these scores as
$\mathbf s_i=(s_{ij})_{j\in\mathcal M_i}$, and collect the scores over all candidate positions as
$\mathbf s_{\mathcal E}=(s_{ij})_{(i,j)\in\mathcal E}$.  A decoder selects
$\mathcal C_i\subseteq\mathcal M_i$ for each sequence, or equivalently the batch-level set
\begin{compactequation}
 \mathcal C=\{(i,j):i\in\{1,\ldots,B\},\ j\in\mathcal C_i\}\subseteq\mathcal E.
 \label{eq:batch-commit-set}
\end{compactequation}
Selected positions are filled with their predicted tokens, while the other positions remain masked.
Thus, under the total budget $T$, the decoder selects $\mathcal C$ to improve the batch's final quality.

\begin{table*}[!t]
\centering
\caption{\textbf{Batch-Paced improves batch-decoding performance over Uniform at a fixed compute budget.}
The table compares both policies on individual benchmarks and mixtures across both
backbones and budget levels.}
\label{tab:main}
\footnotesize
\setlength{\tabcolsep}{0.5pt}
\renewcommand{\arraystretch}{1.10}
\begin{tabular*}{\linewidth}{@{}>{\raggedright\arraybackslash}p{1.53in}@{\hspace{3pt}\extracolsep{\fill}}*{4}{>{\raggedright\arraybackslash}p{3.55em}}>{\columncolor{gray!10}\raggedright\arraybackslash}p{3.55em}*{4}{>{\raggedright\arraybackslash}p{3.55em}}>{\columncolor{gray!10}\raggedright\arraybackslash}p{3.55em}@{}}
\toprule
\multirow{3}{*}{\raisebox{-11pt}{\makebox[1.53in][l]{\hspace{3pt}Benchmark}}} & \multicolumn{5}{c}{LLaDA-8B-Instruct} & \multicolumn{5}{c}{LLaDA 1.5} \\
\cmidrule(lr){2-6}\cmidrule(lr){7-11}
 & \multicolumn{2}{c}{Avg. budget=64} & \multicolumn{2}{c}{Avg. budget=128} & \multicolumn{1}{c}{Full=256} & \multicolumn{2}{c}{Avg. budget=64} & \multicolumn{2}{c}{Avg. budget=128} & \multicolumn{1}{c}{Full=256} \\
\cmidrule(lr){2-3}\cmidrule(lr){4-5}\cmidrule(lr){6-6}\cmidrule(lr){7-8}\cmidrule(lr){9-10}\cmidrule(lr){11-11}
 & \multicolumn{1}{c}{\makebox[3.55em][c]{\shortstack[c]{\fontsize{8.5}{9.2}\selectfont Uniform\\baseline}}} & \multicolumn{1}{c}{\makebox[3.55em][c]{\shortstack[c]{\fontsize{8.5}{9.2}\selectfont Batch-\\Paced}}} & \multicolumn{1}{c}{\makebox[3.55em][c]{\shortstack[c]{\fontsize{8.5}{9.2}\selectfont Uniform\\baseline}}} & \multicolumn{1}{c}{\makebox[3.55em][c]{\shortstack[c]{\fontsize{8.5}{9.2}\selectfont Batch-\\Paced}}} & \multicolumn{1}{c}{\makebox[3.55em][c]{\shortstack[c]{\fontsize{8.5}{9.2}\selectfont Full\\reference}}} & \multicolumn{1}{c}{\makebox[3.55em][c]{\shortstack[c]{\fontsize{8.5}{9.2}\selectfont Uniform\\baseline}}} & \multicolumn{1}{c}{\makebox[3.55em][c]{\shortstack[c]{\fontsize{8.5}{9.2}\selectfont Batch-\\Paced}}} & \multicolumn{1}{c}{\makebox[3.55em][c]{\shortstack[c]{\fontsize{8.5}{9.2}\selectfont Uniform\\baseline}}} & \multicolumn{1}{c}{\makebox[3.55em][c]{\shortstack[c]{\fontsize{8.5}{9.2}\selectfont Batch-\\Paced}}} & \multicolumn{1}{c}{\makebox[3.55em][c]{\shortstack[c]{\fontsize{8.5}{9.2}\selectfont Full\\reference}}} \\
\midrule
\makebox[\linewidth][l]{\hspace{3pt}{\footnotesize GSM8K}} & {\fontsize{8.5}{9.2}\selectfont \ensuremath{67.8_{\scriptscriptstyle\,0.0}}} & {\fontsize{8.5}{9.2}\selectfont \boldmath\ensuremath{77.5_{\scriptscriptstyle\,1.1}^{\ast}}} & {\fontsize{8.5}{9.2}\selectfont \ensuremath{76.2_{\scriptscriptstyle\,0.0}}} & {\fontsize{8.5}{9.2}\selectfont \boldmath\ensuremath{81.0_{\scriptscriptstyle\,0.5}^{\ast}}} & {\fontsize{8.5}{9.2}\selectfont \ensuremath{81.2_{\scriptscriptstyle\,0.0}}} & {\fontsize{8.5}{9.2}\selectfont \ensuremath{70.0_{\scriptscriptstyle\,1.0}}} & {\fontsize{8.5}{9.2}\selectfont \boldmath\ensuremath{80.5_{\scriptscriptstyle\,0.8}^{\ast}}} & {\fontsize{8.5}{9.2}\selectfont \ensuremath{77.7_{\scriptscriptstyle\,0.0}}} & {\fontsize{8.5}{9.2}\selectfont \boldmath\ensuremath{83.8_{\scriptscriptstyle\,0.6}^{\ast}}} & {\fontsize{8.5}{9.2}\selectfont \ensuremath{83.4_{\scriptscriptstyle\,0.0}}} \\
\makebox[\linewidth][l]{\hspace{3pt}{\footnotesize MATH500}} & {\fontsize{8.5}{9.2}\selectfont \ensuremath{25.7_{\scriptscriptstyle\,1.0}}} & {\fontsize{8.5}{9.2}\selectfont \boldmath\ensuremath{32.1_{\scriptscriptstyle\,2.9}^{\ast}}} & {\fontsize{8.5}{9.2}\selectfont \ensuremath{33.2_{\scriptscriptstyle\,1.7}}} & {\fontsize{8.5}{9.2}\selectfont \boldmath\ensuremath{35.5_{\scriptscriptstyle\,2.1}^{\ast}}} & {\fontsize{8.5}{9.2}\selectfont \ensuremath{35.9_{\scriptscriptstyle\,1.7}}} & {\fontsize{8.5}{9.2}\selectfont \ensuremath{27.4_{\scriptscriptstyle\,0.6}}} & {\fontsize{8.5}{9.2}\selectfont \boldmath\ensuremath{32.2_{\scriptscriptstyle\,1.6}^{\ast}}} & {\fontsize{8.5}{9.2}\selectfont \ensuremath{33.1_{\scriptscriptstyle\,0.6}}} & {\fontsize{8.5}{9.2}\selectfont \boldmath\ensuremath{36.4_{\scriptscriptstyle\,1.3}^{\ast}}} & {\fontsize{8.5}{9.2}\selectfont \ensuremath{36.8_{\scriptscriptstyle\,0.8}}} \\
\makebox[\linewidth][l]{\hspace{3pt}{\footnotesize HumanEval}} & {\fontsize{8.5}{9.2}\selectfont \ensuremath{18.4_{\scriptscriptstyle\,0.3}}} & {\fontsize{8.5}{9.2}\selectfont \boldmath\ensuremath{29.3_{\scriptscriptstyle\,1.2}^{\ast}}} & {\fontsize{8.5}{9.2}\selectfont \ensuremath{27.0_{\scriptscriptstyle\,0.3}}} & {\fontsize{8.5}{9.2}\selectfont \boldmath\ensuremath{38.4_{\scriptscriptstyle\,0.4}^{\ast}}} & {\fontsize{8.5}{9.2}\selectfont \ensuremath{41.0_{\scriptscriptstyle\,0.3}}} & {\fontsize{8.5}{9.2}\selectfont \ensuremath{21.3_{\scriptscriptstyle\,0.0}}} & {\fontsize{8.5}{9.2}\selectfont \boldmath\ensuremath{34.6_{\scriptscriptstyle\,0.9}^{\ast}}} & {\fontsize{8.5}{9.2}\selectfont \ensuremath{31.0_{\scriptscriptstyle\,0.3}}} & {\fontsize{8.5}{9.2}\selectfont \boldmath\ensuremath{43.2_{\scriptscriptstyle\,0.9}^{\ast}}} & {\fontsize{8.5}{9.2}\selectfont \ensuremath{45.7_{\scriptscriptstyle\,0.0}}} \\
\makebox[\linewidth][l]{\hspace{3pt}{\footnotesize MBPP}} & {\fontsize{8.5}{9.2}\selectfont \ensuremath{16.9_{\scriptscriptstyle\,0.6}}} & {\fontsize{8.5}{9.2}\selectfont \boldmath\ensuremath{29.3_{\scriptscriptstyle\,1.2}^{\ast}}} & {\fontsize{8.5}{9.2}\selectfont \ensuremath{29.3_{\scriptscriptstyle\,0.8}}} & {\fontsize{8.5}{9.2}\selectfont \boldmath\ensuremath{36.2_{\scriptscriptstyle\,0.6}^{\ast}}} & {\fontsize{8.5}{9.2}\selectfont \ensuremath{37.9_{\scriptscriptstyle\,0.0}}} & {\fontsize{8.5}{9.2}\selectfont \ensuremath{15.9_{\scriptscriptstyle\,1.0}}} & {\fontsize{8.5}{9.2}\selectfont \boldmath\ensuremath{27.0_{\scriptscriptstyle\,1.2}^{\ast}}} & {\fontsize{8.5}{9.2}\selectfont \ensuremath{26.6_{\scriptscriptstyle\,1.2}}} & {\fontsize{8.5}{9.2}\selectfont \boldmath\ensuremath{39.8_{\scriptscriptstyle\,0.9}^{\ast}}} & {\fontsize{8.5}{9.2}\selectfont \ensuremath{41.8_{\scriptscriptstyle\,0.0}}} \\
\makebox[\linewidth][l]{\hspace{3pt}{\footnotesize IFEval}} & {\fontsize{8.5}{9.2}\selectfont \boldmath\ensuremath{44.8_{\scriptscriptstyle\,0.1}}} & {\fontsize{8.5}{9.2}\selectfont \ensuremath{43.8_{\scriptscriptstyle\,1.7}}} & {\fontsize{8.5}{9.2}\selectfont \ensuremath{51.5_{\scriptscriptstyle\,0.1}}} & {\fontsize{8.5}{9.2}\selectfont \boldmath\ensuremath{51.7_{\scriptscriptstyle\,0.8}}} & {\fontsize{8.5}{9.2}\selectfont \ensuremath{55.7_{\scriptscriptstyle\,0.1}}} & {\fontsize{8.5}{9.2}\selectfont \ensuremath{47.2_{\scriptscriptstyle\,0.1}}} & {\fontsize{8.5}{9.2}\selectfont \boldmath\ensuremath{47.7_{\scriptscriptstyle\,0.8}}} & {\fontsize{8.5}{9.2}\selectfont \boldmath\ensuremath{55.2_{\scriptscriptstyle\,0.1}}} & {\fontsize{8.5}{9.2}\selectfont \ensuremath{54.5_{\scriptscriptstyle\,0.7}}} & {\fontsize{8.5}{9.2}\selectfont \ensuremath{58.6_{\scriptscriptstyle\,0.1}}} \\
\midrule
\makebox[\linewidth][l]{\hspace{3pt}{\footnotesize GSM8K+MATH500}} & {\fontsize{8.5}{9.2}\selectfont \ensuremath{46.2_{\scriptscriptstyle\,0.7}}} & {\fontsize{8.5}{9.2}\selectfont \boldmath\ensuremath{55.4_{\scriptscriptstyle\,1.1}^{\ast}}} & {\fontsize{8.5}{9.2}\selectfont \ensuremath{54.2_{\scriptscriptstyle\,0.6}}} & {\fontsize{8.5}{9.2}\selectfont \boldmath\ensuremath{59.0_{\scriptscriptstyle\,1.2}^{\ast}}} & {\fontsize{8.5}{9.2}\selectfont \ensuremath{59.1_{\scriptscriptstyle\,1.1}}} & {\fontsize{8.5}{9.2}\selectfont \ensuremath{49.0_{\scriptscriptstyle\,0.7}}} & {\fontsize{8.5}{9.2}\selectfont \boldmath\ensuremath{56.6_{\scriptscriptstyle\,0.8}^{\ast}}} & {\fontsize{8.5}{9.2}\selectfont \ensuremath{55.5_{\scriptscriptstyle\,0.7}}} & {\fontsize{8.5}{9.2}\selectfont \boldmath\ensuremath{60.2_{\scriptscriptstyle\,1.1}^{\ast}}} & {\fontsize{8.5}{9.2}\selectfont \ensuremath{60.5_{\scriptscriptstyle\,0.5}}} \\
\makebox[\linewidth][l]{\hspace{3pt}{\footnotesize HumanEval+MBPP}} & {\fontsize{8.5}{9.2}\selectfont \ensuremath{16.8_{\scriptscriptstyle\,1.4}}} & {\fontsize{8.5}{9.2}\selectfont \boldmath\ensuremath{31.1_{\scriptscriptstyle\,1.1}^{\ast}}} & {\fontsize{8.5}{9.2}\selectfont \ensuremath{29.5_{\scriptscriptstyle\,1.6}}} & {\fontsize{8.5}{9.2}\selectfont \boldmath\ensuremath{39.1_{\scriptscriptstyle\,0.7}^{\ast}}} & {\fontsize{8.5}{9.2}\selectfont \ensuremath{39.7_{\scriptscriptstyle\,0.3}}} & {\fontsize{8.5}{9.2}\selectfont \ensuremath{18.1_{\scriptscriptstyle\,1.0}}} & {\fontsize{8.5}{9.2}\selectfont \boldmath\ensuremath{30.9_{\scriptscriptstyle\,1.0}^{\ast}}} & {\fontsize{8.5}{9.2}\selectfont \ensuremath{29.8_{\scriptscriptstyle\,0.7}}} & {\fontsize{8.5}{9.2}\selectfont \boldmath\ensuremath{41.6_{\scriptscriptstyle\,0.6}^{\ast}}} & {\fontsize{8.5}{9.2}\selectfont \ensuremath{44.1_{\scriptscriptstyle\,0.5}}} \\
\makebox[\linewidth][l]{\hspace{3pt}{\footnotesize GSM8K+HumanEval}} & {\fontsize{8.5}{9.2}\selectfont \ensuremath{42.7_{\scriptscriptstyle\,0.0}}} & {\fontsize{8.5}{9.2}\selectfont \boldmath\ensuremath{52.8_{\scriptscriptstyle\,1.2}^{\ast}}} & {\fontsize{8.5}{9.2}\selectfont \ensuremath{51.5_{\scriptscriptstyle\,0.0}}} & {\fontsize{8.5}{9.2}\selectfont \boldmath\ensuremath{59.6_{\scriptscriptstyle\,0.7}^{\ast}}} & {\fontsize{8.5}{9.2}\selectfont \ensuremath{60.7_{\scriptscriptstyle\,0.0}}} & {\fontsize{8.5}{9.2}\selectfont \ensuremath{43.6_{\scriptscriptstyle\,0.0}}} & {\fontsize{8.5}{9.2}\selectfont \boldmath\ensuremath{57.1_{\scriptscriptstyle\,1.9}^{\ast}}} & {\fontsize{8.5}{9.2}\selectfont \ensuremath{54.0_{\scriptscriptstyle\,0.0}}} & {\fontsize{8.5}{9.2}\selectfont \boldmath\ensuremath{63.3_{\scriptscriptstyle\,0.5}^{\ast}}} & {\fontsize{8.5}{9.2}\selectfont \ensuremath{64.3_{\scriptscriptstyle\,0.0}}} \\
\makebox[\linewidth][l]{\hspace{3pt}{\footnotesize GSM8K+IFEval}} & {\fontsize{8.5}{9.2}\selectfont \ensuremath{57.2_{\scriptscriptstyle\,0.1}}} & {\fontsize{8.5}{9.2}\selectfont \boldmath\ensuremath{58.2_{\scriptscriptstyle\,1.2}}} & {\fontsize{8.5}{9.2}\selectfont \ensuremath{66.2_{\scriptscriptstyle\,0.1}}} & {\fontsize{8.5}{9.2}\selectfont \boldmath\ensuremath{68.0_{\scriptscriptstyle\,1.0}^{\ast}}} & {\fontsize{8.5}{9.2}\selectfont \ensuremath{69.5_{\scriptscriptstyle\,0.1}}} & {\fontsize{8.5}{9.2}\selectfont \ensuremath{59.1_{\scriptscriptstyle\,0.1}}} & {\fontsize{8.5}{9.2}\selectfont \boldmath\ensuremath{62.9_{\scriptscriptstyle\,0.8}^{\ast}}} & {\fontsize{8.5}{9.2}\selectfont \ensuremath{67.5_{\scriptscriptstyle\,0.7}}} & {\fontsize{8.5}{9.2}\selectfont \boldmath\ensuremath{71.7_{\scriptscriptstyle\,0.4}^{\ast}}} & {\fontsize{8.5}{9.2}\selectfont \ensuremath{72.7_{\scriptscriptstyle\,0.2}}} \\
\makebox[\linewidth][l]{\hspace{3pt}{\footnotesize HumanEval+IFEval}} & {\fontsize{8.5}{9.2}\selectfont \ensuremath{31.0_{\scriptscriptstyle\,0.1}}} & {\fontsize{8.5}{9.2}\selectfont \boldmath\ensuremath{32.7_{\scriptscriptstyle\,2.0}}} & {\fontsize{8.5}{9.2}\selectfont \ensuremath{41.7_{\scriptscriptstyle\,0.1}}} & {\fontsize{8.5}{9.2}\selectfont \boldmath\ensuremath{46.8_{\scriptscriptstyle\,1.3}^{\ast}}} & {\fontsize{8.5}{9.2}\selectfont \ensuremath{49.5_{\scriptscriptstyle\,0.3}}} & {\fontsize{8.5}{9.2}\selectfont \ensuremath{35.3_{\scriptscriptstyle\,0.1}}} & {\fontsize{8.5}{9.2}\selectfont \boldmath\ensuremath{36.3_{\scriptscriptstyle\,1.8}}} & {\fontsize{8.5}{9.2}\selectfont \ensuremath{45.1_{\scriptscriptstyle\,0.1}}} & {\fontsize{8.5}{9.2}\selectfont \boldmath\ensuremath{50.1_{\scriptscriptstyle\,1.3}^{\ast}}} & {\fontsize{8.5}{9.2}\selectfont \ensuremath{53.1_{\scriptscriptstyle\,0.3}}} \\
\midrule
\makebox[\linewidth][l]{\hspace{3pt}{\footnotesize GSM8K+HumanEval+IFEval}} & {\fontsize{8.5}{9.2}\selectfont \ensuremath{43.0_{\scriptscriptstyle\,0.1}}} & {\fontsize{8.5}{9.2}\selectfont \boldmath\ensuremath{48.3_{\scriptscriptstyle\,1.1}^{\ast}}} & {\fontsize{8.5}{9.2}\selectfont \ensuremath{53.3_{\scriptscriptstyle\,0.3}}} & {\fontsize{8.5}{9.2}\selectfont \boldmath\ensuremath{57.8_{\scriptscriptstyle\,0.6}^{\ast}}} & {\fontsize{8.5}{9.2}\selectfont \ensuremath{59.8_{\scriptscriptstyle\,0.1}}} & {\fontsize{8.5}{9.2}\selectfont \ensuremath{45.4_{\scriptscriptstyle\,0.1}}} & {\fontsize{8.5}{9.2}\selectfont \boldmath\ensuremath{51.2_{\scriptscriptstyle\,2.4}^{\ast}}} & {\fontsize{8.5}{9.2}\selectfont \ensuremath{55.7_{\scriptscriptstyle\,0.1}}} & {\fontsize{8.5}{9.2}\selectfont \boldmath\ensuremath{61.9_{\scriptscriptstyle\,0.8}^{\ast}}} & {\fontsize{8.5}{9.2}\selectfont \ensuremath{63.0_{\scriptscriptstyle\,0.2}}} \\
\makebox[\linewidth][l]{\hspace{3pt}{\footnotesize MATH500+MBPP+IFEval}} & {\fontsize{8.5}{9.2}\selectfont \ensuremath{30.2_{\scriptscriptstyle\,1.1}}} & {\fontsize{8.5}{9.2}\selectfont \boldmath\ensuremath{33.8_{\scriptscriptstyle\,2.9}}} & {\fontsize{8.5}{9.2}\selectfont \ensuremath{40.7_{\scriptscriptstyle\,0.5}}} & {\fontsize{8.5}{9.2}\selectfont \boldmath\ensuremath{45.0_{\scriptscriptstyle\,0.8}^{\ast}}} & {\fontsize{8.5}{9.2}\selectfont \ensuremath{46.4_{\scriptscriptstyle\,0.6}}} & {\fontsize{8.5}{9.2}\selectfont \ensuremath{32.0_{\scriptscriptstyle\,1.3}}} & {\fontsize{8.5}{9.2}\selectfont \boldmath\ensuremath{36.3_{\scriptscriptstyle\,1.8}^{\ast}}} & {\fontsize{8.5}{9.2}\selectfont \ensuremath{40.0_{\scriptscriptstyle\,0.3}}} & {\fontsize{8.5}{9.2}\selectfont \boldmath\ensuremath{46.0_{\scriptscriptstyle\,1.5}^{\ast}}} & {\fontsize{8.5}{9.2}\selectfont \ensuremath{48.4_{\scriptscriptstyle\,0.3}}} \\
\makebox[\linewidth][l]{\hspace{3pt}{\footnotesize GSM8K+MATH500+MBPP}} & {\fontsize{8.5}{9.2}\selectfont \ensuremath{36.1_{\scriptscriptstyle\,1.1}}} & {\fontsize{8.5}{9.2}\selectfont \boldmath\ensuremath{46.8_{\scriptscriptstyle\,1.2}^{\ast}}} & {\fontsize{8.5}{9.2}\selectfont \ensuremath{47.0_{\scriptscriptstyle\,0.8}}} & {\fontsize{8.5}{9.2}\selectfont \boldmath\ensuremath{51.8_{\scriptscriptstyle\,1.3}^{\ast}}} & {\fontsize{8.5}{9.2}\selectfont \ensuremath{52.4_{\scriptscriptstyle\,1.0}}} & {\fontsize{8.5}{9.2}\selectfont \ensuremath{38.4_{\scriptscriptstyle\,1.1}}} & {\fontsize{8.5}{9.2}\selectfont \boldmath\ensuremath{46.6_{\scriptscriptstyle\,1.1}^{\ast}}} & {\fontsize{8.5}{9.2}\selectfont \ensuremath{46.6_{\scriptscriptstyle\,1.2}}} & {\fontsize{8.5}{9.2}\selectfont \boldmath\ensuremath{54.2_{\scriptscriptstyle\,0.9}^{\ast}}} & {\fontsize{8.5}{9.2}\selectfont \ensuremath{55.2_{\scriptscriptstyle\,0.9}}} \\
\makebox[\linewidth][l]{\hspace{3pt}{\footnotesize GSM8K+MATH500+IFEval}} & {\fontsize{8.5}{9.2}\selectfont \ensuremath{46.5_{\scriptscriptstyle\,0.9}}} & {\fontsize{8.5}{9.2}\selectfont \boldmath\ensuremath{49.6_{\scriptscriptstyle\,1.1}^{\ast}}} & {\fontsize{8.5}{9.2}\selectfont \ensuremath{56.2_{\scriptscriptstyle\,0.6}}} & {\fontsize{8.5}{9.2}\selectfont \boldmath\ensuremath{59.0_{\scriptscriptstyle\,0.6}^{\ast}}} & {\fontsize{8.5}{9.2}\selectfont \ensuremath{60.0_{\scriptscriptstyle\,0.8}}} & {\fontsize{8.5}{9.2}\selectfont \ensuremath{49.2_{\scriptscriptstyle\,1.4}}} & {\fontsize{8.5}{9.2}\selectfont \boldmath\ensuremath{52.8_{\scriptscriptstyle\,1.1}^{\ast}}} & {\fontsize{8.5}{9.2}\selectfont \ensuremath{57.3_{\scriptscriptstyle\,1.0}}} & {\fontsize{8.5}{9.2}\selectfont \boldmath\ensuremath{60.7_{\scriptscriptstyle\,0.9}^{\ast}}} & {\fontsize{8.5}{9.2}\selectfont \ensuremath{61.9_{\scriptscriptstyle\,0.4}}} \\
\makebox[\linewidth][l]{\hspace{3pt}{\footnotesize HumanEval+MBPP+IFEval}} & {\fontsize{8.5}{9.2}\selectfont \ensuremath{26.1_{\scriptscriptstyle\,1.0}}} & {\fontsize{8.5}{9.2}\selectfont \boldmath\ensuremath{33.1_{\scriptscriptstyle\,1.3}^{\ast}}} & {\fontsize{8.5}{9.2}\selectfont \ensuremath{38.7_{\scriptscriptstyle\,0.8}}} & {\fontsize{8.5}{9.2}\selectfont \boldmath\ensuremath{45.2_{\scriptscriptstyle\,0.6}^{\ast}}} & {\fontsize{8.5}{9.2}\selectfont \ensuremath{46.3_{\scriptscriptstyle\,0.1}}} & {\fontsize{8.5}{9.2}\selectfont \ensuremath{29.2_{\scriptscriptstyle\,1.6}}} & {\fontsize{8.5}{9.2}\selectfont \boldmath\ensuremath{34.3_{\scriptscriptstyle\,0.9}^{\ast}}} & {\fontsize{8.5}{9.2}\selectfont \ensuremath{39.1_{\scriptscriptstyle\,0.5}}} & {\fontsize{8.5}{9.2}\selectfont \boldmath\ensuremath{46.9_{\scriptscriptstyle\,0.8}^{\ast}}} & {\fontsize{8.5}{9.2}\selectfont \ensuremath{49.4_{\scriptscriptstyle\,0.2}}} \\
\makebox[\linewidth][l]{\hspace{3pt}{\footnotesize GSM8K+HumanEval+MBPP}} & {\fontsize{8.5}{9.2}\selectfont \ensuremath{33.3_{\scriptscriptstyle\,0.8}}} & {\fontsize{8.5}{9.2}\selectfont \boldmath\ensuremath{47.0_{\scriptscriptstyle\,0.5}^{\ast}}} & {\fontsize{8.5}{9.2}\selectfont \ensuremath{45.0_{\scriptscriptstyle\,0.9}}} & {\fontsize{8.5}{9.2}\selectfont \boldmath\ensuremath{51.8_{\scriptscriptstyle\,0.6}^{\ast}}} & {\fontsize{8.5}{9.2}\selectfont \ensuremath{52.5_{\scriptscriptstyle\,0.1}}} & {\fontsize{8.5}{9.2}\selectfont \ensuremath{34.3_{\scriptscriptstyle\,1.0}}} & {\fontsize{8.5}{9.2}\selectfont \boldmath\ensuremath{48.4_{\scriptscriptstyle\,0.8}^{\ast}}} & {\fontsize{8.5}{9.2}\selectfont \ensuremath{45.7_{\scriptscriptstyle\,0.6}}} & {\fontsize{8.5}{9.2}\selectfont \boldmath\ensuremath{55.4_{\scriptscriptstyle\,0.3}^{\ast}}} & {\fontsize{8.5}{9.2}\selectfont \ensuremath{56.8_{\scriptscriptstyle\,0.6}}} \\
\bottomrule
\end{tabular*}

\end{table*}

\section{Method}
\label{sec:method}

\subsection{Independent Sequence-Wise Allocation}
\label{sec:sequence-wise-allocation}

Independent Sequence-Wise Allocation allocates the budget independently to each sequence.

\paragraph{Uniform.}
Uniform commits the same $r=\lceil L/\bar T\rceil$ highest-scoring positions from each sequence's
$\mathcal M_i$ according to $\mathbf s_i$:
 \begin{compactequation}
 \mathcal C^{\mathrm{uni}}
 =\bigcup_{i=1}^{B}\{(i,j):j\in\operatorname{Top}_{r}(\mathcal M_i;\mathbf s_i)\}.
 \label{eq:uniform-policy}
 \end{compactequation}

\paragraph{Fast-dLLM.}
Fast-dLLM~\cite{wu2026fast} uses the same confidence threshold $\theta$ for every sequence and commits
the batch-level set
\begin{compactequation}
 \mathcal C^{\mathrm{fast}}(\theta)=
 \bigcup_{i=1}^{B}\{(i,j):j\in\mathcal M_i,\ s_{ij}\geq\theta\}.
 \label{eq:fast-policy}
\end{compactequation}
Different thresholds induce different numbers of denoising forward passes.  The matched-budget
threshold-selection protocol is described with the Fast-dLLM comparison in Section~\ref{sec:fastdllm}.

\subsection{Global Compute Allocation within a Batch}
\label{sec:batch-level-allocation}

We introduce Batch-Paced, which jointly optimizes decoding-compute allocation across sequences by
ranking masked-token confidence across the batch, allowing different sequences to advance at different
paces. All sequences draw from a shared budget pool, so unused budget from sequences
that finish early can be transferred to sequences that continue decoding.  By controlling the number
of committed tokens at each pass, Batch-Paced controls how the fixed total budget is allocated.

Specifically, at each denoising pass, define the unfinished sequences and their candidate sets as
$\mathcal A=\{i:\mathcal M_i\neq\varnothing\}$,
$\mathcal E_i=\{(i,j):j\in\mathcal M_i\}$, and
$\mathcal E=\bigcup_{i\in\mathcal A}\mathcal E_i$.  A batched forward pass costs
$|\mathcal A|$ units; the shared pool is initialized as $R=B\bar T$ and decreases by
$|\mathcal A|$ after each pass.  When a sequence finishes, it leaves $\mathcal A$, reducing the cost
of later passes and leaving more of $R$ for the remaining sequences.  Let
$U=\sum_{i\in\mathcal A}|\mathcal M_i|$.  Batch-Paced uses the remaining pool to determine $K$,
the number of positions committed at this pass:
\begin{compactequation}
 K=\min\left\{|\mathcal E|,
       \left\lceil\frac{U}{\max(1,R/|\mathcal A|)}\right\rceil\right\}.
 \label{eq:batch-paced-k}
\end{compactequation}
It then selects the corresponding positions globally according to their confidence scores:
\begin{compactequation}
 \mathcal C^{\mathrm{BP}}=\operatorname{Top}_{K}(\mathcal E;\mathbf s_{\mathcal E}).
 \label{eq:batch-paced-policy}
\end{compactequation}
The selection enforces a minimum commit of one position for every unfinished sequence: if the pooled
top-$K$ candidates omit sequence $i$, its highest-scoring position in $\mathcal M_i$ is included.
For sequence $i$, the committed positions are
$\mathcal C_i^{\mathrm{BP}}=\{j:(i,j)\in\mathcal C^{\mathrm{BP}}\}$, whose size may differ across
sequences.  Thus, Batch-Paced performs batch-level decoding under a fixed total budget by jointly
selecting commit positions across sequences. Algorithm~\ref{alg:batch-paced} summarizes this base
per-pass selector; Appendix~\ref{sec:round-capped} extends Batch-Paced with an explicit batch-round
allowance and gives the resulting full decoding algorithm in Algorithm~\ref{alg:round-capped}.

\begin{figure*}[!t]
\centering
\includegraphics[width=0.99\textwidth]{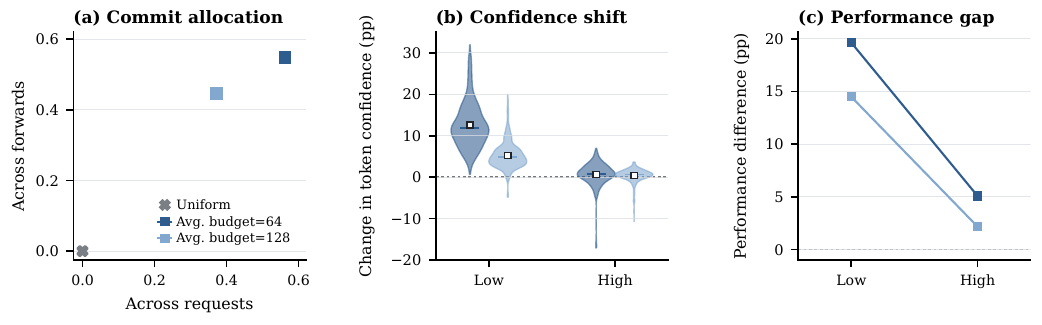}
\caption{\textbf{Batch-Paced's largest confidence shifts and performance gains occur in the low-confidence group.}
The panels show (a) variation in commit allocation across requests and forwards, (b) confidence shifts by
Uniform-confidence group, and (c) paired performance differences between Batch-Paced and Uniform.}
\label{fig:allocation-evidence}
\end{figure*}

\begin{figure*}[!t]
\centering
\includegraphics[width=0.99\textwidth]{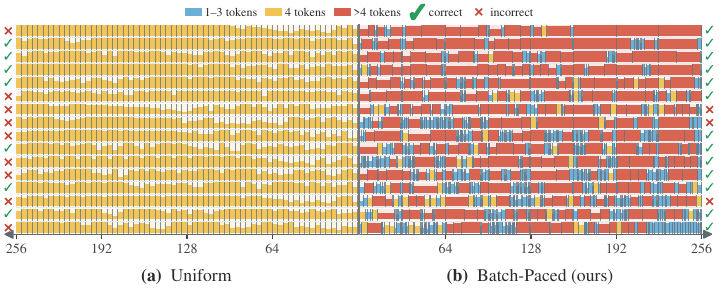}
\caption{\textbf{Batch-Paced gives requests different commit schedules within a fixed-budget batch.}
Matched panels show the same illustrative mixed $B=16$ batch with requests aligned row by row; the
shared center separates Uniform's leftward and Batch-Paced's rightward token-progress axes. Segment
width, height, color, and row-end marks encode per-forward token count, mean confidence, commit size,
and final correctness, respectively.}
\label{fig:mechanism}
\end{figure*}

\begin{algorithm}[t]
\caption{Batch-Paced per-pass selection}
\label{alg:batch-paced}
\begin{algorithmic}[1]
\Require unfinished sequences $\mathcal A$, candidate sets $\{\mathcal E_i\}$, scores $s_{ij}$, remaining pool $R\geq |\mathcal A|$
\State $\mathcal E\gets\bigcup_{i\in\mathcal A}\mathcal E_i$; compute $K$ using Eq.~\ref{eq:batch-paced-k}
\State $\mathcal C\gets\operatorname{Top}_{K}(\mathcal E; s_{ij})$
\ForAll{$i\in\mathcal A$}
  \If{$|\mathcal C\cap\mathcal E_i|=0$}
    \State $\mathcal C\gets\mathcal C\cup\{\arg\max_{(i,j)\in\mathcal E_i}s_{ij}\}$
  \EndIf
\EndFor
\State Commit positions in $\mathcal C$; advance any completed sequence
\State $R\gets R-|\mathcal A|$
\State \Return updated batch and pool
\end{algorithmic}
\end{algorithm}

\section{Experiments}
\label{sec:experiments}

\subsection{Experimental Setup}
\label{sec:setup}

We evaluate GSM8K~\citep{cobbe2021training}, MATH500~\citep{hendrycks2021measuring,lightman2024let},
HumanEval~\citep{chen2021evaluating}, MBPP~\citep{austin2021program}, and IFEval~\citep{zhou2023instruction}
individually and in cross-domain mixtures, reflecting serving batches that may contain requests from different
task domains. For each benchmark, we fix a pool of up to 256 examples; each mixture combines equal-sized
samples from its component pools. We shuffle the examples and partition them into batches of $B=16$,
allowing a smaller final batch when needed. Across five batch-formation seeds, the example pool remains fixed
while its order and batch membership vary. We compute accuracy over the full example set for each arrangement
and report the mean and standard deviation across arrangements, with standard deviations shown as table
subscripts. Across tables, boldface marks the higher mean in each comparison, regardless of significance; an
asterisk marks a nominally significant difference under a two-sided paired $t$-test over five matched seeds
($p<0.05$). Prompts and exemplars follow the corresponding lm-evaluation-harness task definitions where
available~\citep{eval-harness}. We evaluate LLaDA-8B-Instruct and LLaDA 1.5~\citep{nie2026large,zhu2026llada};
both use the same semi-autoregressive remasking setup as prior LLaDA evaluations~\citep{nie2026large,zhu2026llada},
with generation length $L=256$, block length $H=32$, and top-1 softmax probability as the confidence score.

\subsection{Main Results}
\label{sec:main-results}

\begin{table*}[!t]
\centering
\footnotesize
\setlength{\tabcolsep}{4pt}
\renewcommand{\arraystretch}{1.0}
\caption{\textbf{Batch-Paced largely retains its performance advantage across decoding configurations.}
The table compares Uniform and Batch-Paced on GSM8K--HumanEval across both models and two budget fractions, varying block and generation lengths and confidence-score functions; entries list Uniform/Batch-Paced mean accuracy.}
\label{tab:sensitivity-summary}
\begin{tabular*}{\textwidth}{@{\extracolsep{\fill}}ll*{4}{c}@{}}
\toprule
\multirow{2}{*}{Factor} & \multirow{2}{*}{Setting} &
\multicolumn{2}{c}{\shortstack{LLaDA-8B-Instruct\\Avg. budget / gen. length}} &
\multicolumn{2}{c}{\shortstack{LLaDA 1.5\\Avg. budget / gen. length}} \\
\cmidrule(lr){3-4} \cmidrule(lr){5-6}
& & $1/4$ & $1/2$ & $1/4$ & $1/2$ \\
\midrule
\multirow{2}{*}{Block length} & $H=64$ & $43.0_{0.0}/\mathbf{53.8}_{1.5}$ & $52.1_{0.0}/\mathbf{58.1}_{1.0}$ & $47.0_{0.0}/\mathbf{58.7}_{1.4}$ & $55.1_{1.5}/\mathbf{62.4}_{0.7}$ \\
& $H=256$ & $43.3_{0.0}/\mathbf{46.6}_{1.5}$ & $\mathbf{50.0}_{0.0}/48.3_{0.4}$ & $40.7_{0.4}/\mathbf{48.8}_{1.1}$ & $49.1_{0.0}/\mathbf{52.8}_{0.5}$ \\
\midrule
\multirow{2}{*}{Generation length} & $L=128$ & $43.0_{0.0}/\mathbf{50.9}_{2.1}$ & $52.4_{0.0}/\mathbf{56.5}_{0.4}$ & $43.6_{0.0}/\mathbf{52.1}_{1.2}$ & $53.4_{0.0}/\mathbf{57.7}_{0.3}$ \\
& $L=512$ & $44.0_{1.2}/\mathbf{59.1}_{2.2}$ & $57.1_{1.2}/\mathbf{61.3}_{0.8}$ & $45.0_{0.9}/\mathbf{59.6}_{2.1}$ & $60.5_{0.3}/\mathbf{63.4}_{0.8}$ \\
\midrule
\multirow{2}{*}{\shortstack{Confidence-score\\function}} & Top-1/2 margin & $45.2_{1.5}/\mathbf{57.3}_{1.5}^{*}$ & $53.4_{0.0}/\mathbf{59.6}_{0.4}^{*}$ & $45.4_{0.0}/\mathbf{57.6}_{1.7}^{*}$ & $57.3_{0.0}/\mathbf{61.7}_{0.4}^{*}$ \\
& Negative entropy & $34.2_{0.8}/\mathbf{47.0}_{2.2}^{*}$ & $51.2_{0.0}/\mathbf{59.4}_{0.5}^{*}$ & $40.5_{0.0}/\mathbf{48.7}_{1.8}^{*}$ & $53.0_{0.0}/\mathbf{62.3}_{0.8}^{*}$ \\
\bottomrule
\end{tabular*}
\end{table*}

\begin{figure}[!t]
\centering
\includegraphics[width=\columnwidth]{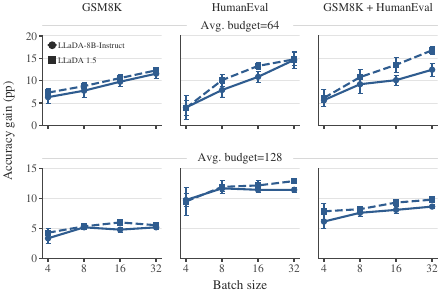}
\caption{\textbf{Batch-Paced's performance advantage over Uniform generally grows with batch size.}
Curves show paired accuracy gains across single-task and mixed-task settings for both
models at both budgets.}
\label{fig:batchsize-analysis}
\end{figure}

Table~\ref{tab:main} reports individual benchmarks and representative task mixtures.
The mixtures cover both benchmarks from the same domain, such as GSM8K with MATH500 and HumanEval
with MBPP, and batches that combine different domains, such as GSM8K with HumanEval, GSM8K with
IFEval, and HumanEval with IFEval. Three-benchmark mixtures broaden the task composition. For each
backbone, Uniform and \ours{} are compared at average
denoising budgets of 64 and 128. At full budget (256), each sequence commits one token per
forward, so we report one reference.
Across both backbones, \ours{} improves on most individual benchmarks at both reduced budgets and
on every reported multi-benchmark mixture. Gains are especially pronounced on code benchmarks and
code-containing mixtures, yet positive for within- and cross-domain mixtures.

\begin{table*}[!t]
\centering
\caption{\textbf{Fast-dLLM and Batch-Paced are broadly comparable.}
The table compares their accuracy on individual benchmarks and representative mixtures across both
backbones and two target denoising budgets.}
\label{tab:fast-secondary}
\footnotesize
\setlength{\tabcolsep}{0.5pt}
\renewcommand{\arraystretch}{1.08}
\begin{tabular*}{\linewidth}{@{}>{\raggedright\arraybackslash}p{1.72in}@{\hspace{3pt}\extracolsep{\fill}}*{8}{>{\centering\arraybackslash}p{0.43in}}@{}}
\toprule
\multirow{3}{*}{\raisebox{-11pt}{\makebox[1.72in][l]{\hspace{3pt}Benchmark}}} & \multicolumn{4}{c}{LLaDA-8B-Instruct} & \multicolumn{4}{c}{LLaDA 1.5} \\
\cmidrule(lr){2-5}\cmidrule(lr){6-9}
 & \multicolumn{2}{c}{Avg. budget=64} & \multicolumn{2}{c}{Avg. budget=128} & \multicolumn{2}{c}{Avg. budget=64} & \multicolumn{2}{c}{Avg. budget=128} \\
\cmidrule(lr){2-3}\cmidrule(lr){4-5}\cmidrule(lr){6-7}\cmidrule(lr){8-9}
 & \shortstack[c]{Fast-\\dLLM} & \shortstack[c]{Batch-\\Paced} & \shortstack[c]{Fast-\\dLLM} & \shortstack[c]{Batch-\\Paced} & \shortstack[c]{Fast-\\dLLM} & \shortstack[c]{Batch-\\Paced} & \shortstack[c]{Fast-\\dLLM} & \shortstack[c]{Batch-\\Paced} \\
\midrule
HumanEval & {\boldmath$36.6_{\scriptscriptstyle\,1.9}^{*}$} & $29.3_{\scriptscriptstyle\,1.2}$ & {\boldmath$40.9_{\scriptscriptstyle\,0.0}^{*}$} & $38.4_{\scriptscriptstyle\,0.4}$ & $33.0_{\scriptscriptstyle\,11.5}$ & {\boldmath$34.6_{\scriptscriptstyle\,0.9}$} & {\boldmath$45.7_{\scriptscriptstyle\,0.0}^{*}$} & $43.2_{\scriptscriptstyle\,0.9}$ \\
IFEval & $39.6_{\scriptscriptstyle\,1.9}$ & {\boldmath$43.8_{\scriptscriptstyle\,1.7}^{*}$} & $50.5_{\scriptscriptstyle\,1.9}$ & {\boldmath$51.7_{\scriptscriptstyle\,0.8}$} & $40.4_{\scriptscriptstyle\,1.6}$ & {\boldmath$47.7_{\scriptscriptstyle\,0.8}^{*}$} & $51.7_{\scriptscriptstyle\,0.7}$ & {\boldmath$54.5_{\scriptscriptstyle\,0.7}^{*}$} \\
GSM8K & {\boldmath$78.8_{\scriptscriptstyle\,0.4}$} & $77.5_{\scriptscriptstyle\,1.1}$ & {\boldmath$81.2_{\scriptscriptstyle\,0.0}$} & $81.0_{\scriptscriptstyle\,0.5}$ & {\boldmath$81.6_{\scriptscriptstyle\,0.9}$} & $80.5_{\scriptscriptstyle\,0.8}$ & $83.3_{\scriptscriptstyle\,0.1}$ & {\boldmath$83.8_{\scriptscriptstyle\,0.6}$} \\
\cmidrule(lr){1-9}
GSM8K + IFEval & $46.0_{\scriptscriptstyle\,2.7}$ & {\boldmath$58.2_{\scriptscriptstyle\,1.2}^{*}$} & {\boldmath$68.8_{\scriptscriptstyle\,0.7}$} & $68.3_{\scriptscriptstyle\,1.0}$ & $45.0_{\scriptscriptstyle\,3.7}$ & {\boldmath$62.9_{\scriptscriptstyle\,0.8}^{*}$} & $71.2_{\scriptscriptstyle\,0.9}$ & {\boldmath$71.7_{\scriptscriptstyle\,0.4}$} \\
GSM8K + HumanEval & {\boldmath$57.9_{\scriptscriptstyle\,0.1}^{*}$} & $52.8_{\scriptscriptstyle\,1.2}$ & {\boldmath$60.7_{\scriptscriptstyle\,0.0}^{*}$} & $59.6_{\scriptscriptstyle\,0.7}$ & {\boldmath$60.6_{\scriptscriptstyle\,1.4}^{*}$} & $57.1_{\scriptscriptstyle\,1.9}$ & {\boldmath$64.3_{\scriptscriptstyle\,0.1}^{*}$} & $63.3_{\scriptscriptstyle\,0.5}$ \\
HumanEval + IFEval & $24.0_{\scriptscriptstyle\,2.8}$ & {\boldmath$32.7_{\scriptscriptstyle\,2.0}^{*}$} & $46.2_{\scriptscriptstyle\,3.0}$ & {\boldmath$46.8_{\scriptscriptstyle\,1.3}$} & $27.2_{\scriptscriptstyle\,4.3}$ & {\boldmath$36.3_{\scriptscriptstyle\,1.8}^{*}$} & $48.5_{\scriptscriptstyle\,5.7}$ & {\boldmath$50.1_{\scriptscriptstyle\,1.3}$} \\
\cmidrule(lr){1-9}
GSM8K + HumanEval + IFEval & $33.9_{\scriptscriptstyle\,5.6}$ & {\boldmath$48.3_{\scriptscriptstyle\,1.1}^{*}$} & {\boldmath$59.3_{\scriptscriptstyle\,1.0}$} & $57.3_{\scriptscriptstyle\,0.9}$ & $36.6_{\scriptscriptstyle\,2.3}$ & {\boldmath$51.2_{\scriptscriptstyle\,2.4}^{*}$} & $61.8_{\scriptscriptstyle\,1.2}$ & {\boldmath$61.9_{\scriptscriptstyle\,0.8}$} \\
\bottomrule
\end{tabular*}

\end{table*}

\subsection{Batch-Paced Reshapes Decoding}
\label{sec:allocation-dynamics}

Batch-Paced shapes each request's decoding dynamics by considering token confidence scores jointly
across the batch. Next, we ask whether this joint consideration changes
denoising pace across requests, how the resulting pace reshapes the confidence-score
trajectory of committed tokens over successive steps, and whether these shifts are reflected in
final decoding performance. Figure~\ref{fig:allocation-evidence} traces these dynamics on the
GSM8K--HumanEval mixture for LLaDA 1.5; Appendix~\ref{app:allocation-dynamics-8b} presents the
corresponding analysis for LLaDA-8B-Instruct.

Panel (a) quantifies commit-allocation unevenness with normalized Gini coefficients across requests
within each forward and across forwards within each request; higher values indicate less even
allocation, and Uniform is zero by construction. Batch-Paced varies allocation along both dimensions.
Figure~\ref{fig:mechanism} illustrates this redistribution in a matched batch: Uniform assigns fixed
commit counts per request, whereas Batch-Paced produces varying request-level schedules. Requests are grouped by their mean
committed-token confidence under Uniform. In (b), Batch-Paced yields larger increases in mean
committed-token confidence for the lower-confidence group, with smaller shifts for the higher-confidence
group. In (c), the lower-confidence group shows larger paired performance gains, while the
higher-confidence group also improves. Together, Batch-Paced's changed denoising pace accompanies
higher committed-token confidence, especially for lower-confidence requests, and better final decoding
performance.

\subsection{Batch-Size Scaling}
\label{sec:batchsize-analysis}

In batched dLLM serving, requests are decoded together, and the batch size determines how many
sequences participate in each batched decoding step.  This raises a practical question about
batched decoding: as more requests share a batch, can decoding the batch as a whole improve task
performance under the same average denoising budget per request?  We compare Batch-Paced with Uniform at
$B\in\{4,8,16,32\}$ for both backbones, three task settings, and two average per-sequence budgets.
Within each comparison, we hold $H=32$, $L=256$, the confidence score, task setting, and average
per-sequence budget fixed, so increasing $B$ expands the set of requests considered together while
preserving each request's compute allowance.  Figure~\ref{fig:batchsize-analysis} shows the paired
accuracy gain over Uniform as batch size varies.  Batch-Paced's gains generally increase with $B$ across
both backbones, all three task settings, and
both budgets.  At $B=32$, its mean gain exceeds that at $B=4$ in every condition, with especially
clear growth at $\bar T=64$.  Larger batches may give joint allocation more choices than Uniform's
sequence-wise schedule.

\subsection{Budget and Decoding Robustness}
\label{sec:sensitivity}

On the 1:1 GSM8K--HumanEval mixture, we test changes to block and generation length and two alternative
confidence-score functions: the top-1/top-2 probability margin, a confidence-gap signal used by Prophet~\citep{li2026diffusion},
and negative entropy, related to EB-Sampler's entropy-based token ordering~\citep{ben2026accelerated}.
Table~\ref{tab:sensitivity-summary} compares Uniform and Batch-Paced across these settings for both models and
two budget fractions. Batch-Paced largely retains its performance advantage across these decoding configurations.
Appendix~\ref{app:decoding-choices} reports the budget sweep (Figure~\ref{fig:budget-sensitivity})
and trends across block and generation lengths (Figures~\ref{fig:block-sensitivity} and
\ref{fig:genlength-sensitivity}); Appendix~\ref{sec:ratio-sensitivity} covers benchmark proportions.

\subsection{Comparison with Adaptive Decoding under Search-Based Budget Matching}
\label{sec:fastdllm}

Recent dLLM decoders use signals such as token confidence, entropy, temporal consistency, and
denoising history to adapt token commitments within individual sequences
\citep{wu2026fast,ben2026accelerated,kim2026klass,wang2026creditdecoding}. Fast-dLLM is a strong
representative adaptive decoder that applies a confidence threshold independently to each
sequence, yielding different commit paces without specifying a batch budget; its compute varies by model,
task, and batch mix.

\begin{figure*}[!t]
\centering
\includegraphics[width=0.99\textwidth]{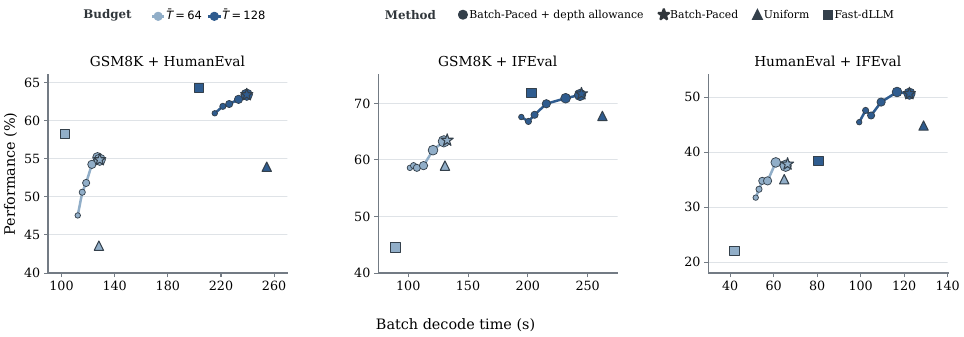}
\caption{\textbf{Depth limits trade Batch-Paced performance for shorter decode time.}
The plot compares Uniform, Fast-dLLM, and original and depth-limited Batch-Paced across task mixtures
and budgets; larger circles denote larger depth allowances. Batch-Paced generally outperforms Uniform
at comparable decode time; tighter allowances move it toward Fast-DLLM's runtime; larger ones recover
the original policy's performance and runtime.}
\label{fig:round-allowance}
\end{figure*}

An explicit batch budget is useful when a serving system must limit the denoising work assigned to
each batch. For example, a shared-GPU scheduler can set per-batch allowances to divide compute across
batches, while a service can offer different compute tiers by assigning different allowances. We
compare Batch-Paced with Fast-dLLM at matched batch budgets to test whether joint batch decisions
achieve comparable performance. For each model, task setting, budget, and seed, we use at most eight
bisection steps to select the tested $\theta$ whose largest per-batch average denoising budget per
request is closest to $\bar T$, while keeping every evaluation batch within the target. Probe-decode
compute is excluded. Table~\ref{tab:fast-secondary} shows broadly comparable performance, with each
method leading in some settings. To isolate budget stopping from joint token selection, we add two
Fast-dLLM stopping rules that retain its threshold-based commits: one force-completes a sequence
when its individual allowance is exhausted; the other force-completes unfinished sequences when the
shared batch budget cannot fund another pass. Batch-Paced generally achieves higher mean performance
on benchmark mixtures; Appendix~\ref{sec:allocation-variants} details both variants, and
Table~\ref{tab:allocation-variants} reports results.

\subsection{Batch Decode-Time and Performance Trade-offs with a Batch-Paced Extension}
\label{sec:depth-tradeoff}

Batch-Paced changes each request's denoising pace to improve overall batch performance, but can
require more sequential denoising steps and thus increase batch decode time. Since completion time
matters alongside output quality in batched serving, we evaluate both. To explore this trade-off, we
extend Batch-Paced with a decoding-depth allowance $D_{\max}$ that limits the number of batched
denoising steps. The original policy commits at least one token per unfinished request at each step;
under the allowance, this minimum rises as the remaining depth shrinks, while additional commits still
follow the batch-wide confidence ranking. Appendix~\ref{sec:round-capped} gives the full algorithm.

We evaluate three pairwise cross-domain mixtures on both backbones at target budgets
$\bar T\in\{64,128\}$, using the same sampled examples and batch order across policies. For each
setting, we test $D_{\max}/\bar T=1$ and $1+2^{-n}$ for $n\in\{0,1,2,3,4\}$, spanning added
allowances from $\bar T/16$ to $\bar T$. Figure~\ref{fig:round-allowance} shows LLaDA 1.5;
Appendix~\ref{app:round-allowance-8b} shows LLaDA-8B-Instruct results in Figure~\ref{fig:round-allowance-8b}.
For timing, all policies use the same Transformers/PyTorch implementation on a single NVIDIA
A100-SXM4-80GB GPU, with serving-engine acceleration disabled. Figure~\ref{fig:round-allowance}
compares batch performance and decode time for the depth-limited variants, Uniform, Fast-dLLM, and
original Batch-Paced. Batch-Paced generally outperforms Uniform at broadly comparable decode time,
running slightly slower under tighter budgets and slightly faster under larger ones. Fast-dLLM is
faster across budgets, but the performance lead varies:
Batch-Paced leads more often under tighter budgets, and Fast-dLLM under larger ones.
Tightening $D_{\max}$ shortens Batch-Paced decoding toward Fast-dLLM's runtime, at some performance
cost; increasing it moves performance and decode time back toward the original policy. Together,
these results show that Batch-Paced improves performance over Uniform at comparable decode time,
while its depth allowance can trade modest performance for runtime closer to Fast-dLLM, a strong
adaptive decoder that requires setting-specific threshold search to meet a target compute budget.

\section{Conclusion}

Serving systems may bound each batch's compute, leaving its allocation across requests open. We
formulate this fixed-budget dLLM batch-decoding problem and introduce Batch-Paced, a training-free
method that jointly ranks masked-token confidence across the batch. Across two backbones and five
benchmarks with mixtures, Batch-Paced generally improves matched-budget performance over Uniform's
independent sequence-wise allocation. Gains tend to grow with batch size and persist across tested
decoding configurations. Analysis of decoding dynamics indicates that the largest increases in
committed-token confidence and performance occur among lower-confidence requests. At matched budgets,
Batch-Paced performs broadly comparably to Fast-dLLM, a representative sequence-wise adaptive decoder
whose confidence threshold requires setting-specific calibration. Its batch decode time is comparable
to Uniform's but generally longer than Fast-dLLM's. A decoding-depth allowance narrows this gap at some
performance cost, yielding a tunable balance between batch quality and batch decode time.

\section*{Limitations}

First, our evaluation is limited to mask-to-token decoding and does not cover more flexible dLLM
decoding schemes, including arbitrary-order and infilling generation~\citep{ye2508dream} or editing
previously committed tokens~\citep{bie2026llada2}. Second, our workload evaluation is limited to standard benchmarks
and constructed benchmark mixtures rather than requests sampled from deployed services; these may
not fully reflect the diversity and task proportions of real-world requests. Finally, our timing
evaluation is limited to synchronized decode time measured for each pre-formed evaluation batch and
averaged across batches; it does not capture end-to-end latency in a continuous online request stream,
where request arrivals and dynamic batch formation also matter.
Despite these limitations, to our knowledge, this is the first controlled study of allocating a fixed
denoising budget across requests to improve dLLM batch quality. By using benchmark-derived batches as a
tractable proxy for heterogeneous request groups, we provide initial insight into this underexplored
problem and a basis for future evaluation on live serving traces and more diverse decoding schemes.

\bibliography{custom}

@inproceedings{arriola2025block,
  title={Block diffusion: Interpolating between autoregressive and diffusion language models},
  author={Arriola, Marianne and Gokaslan, Aaron and Chiu, Justin and Yang, Zhihan and Qi, Zhixuan and Han, Jiaqi and Sahoo, Subham and Kuleshov, Volodymyr},
  booktitle={International Conference on Learning Representations},
  volume={2025},
  pages={50726--50753},
  year={2025}
}

@article{bie2025llada2,
  title={Llada2. 0: Scaling up diffusion language models to 100b},
  author={Bie, Tiwei and Cao, Maosong and Chen, Kun and Du, Lun and Gong, Mingliang and Gong, Zhuochen and Gu, Yanmei and Hu, Jiaqi and Huang, Zenan and Lan, Zhenzhong and others},
  journal={arXiv preprint arXiv:2512.15745},
  year={2025}
}

@article{team2026diffusiongemma,
  title={DiffusionGemma technical report},
  author={Team, DiffusionGemma and Ta{\"\i}ga, Adrien Ali and Assiene, James and Calandriello, Daniele and Chaabouni, Rahma and Gante, Jo{\~a}o and von Glehn, Tamara and Keating, Nate and Knutsen, Chris and Kukla, Martin and others},
  journal={arXiv preprint arXiv:2608.00146},
  year={2026}
}

@misc{anthropiclimits,
  author       = {{Anthropic}},
  title        = {Our Approach to API Rate Limits},
  howpublished = {Anthropic Help Center},
  year         = {2026},
  note         = {Accessed October 9, 2026},
  url          = {https://support.anthropic.com/en/articles/8243635-our-approach-to-api-rate-limits}
}

@article{ben2026accelerated,
  title={Accelerated sampling from masked diffusion models via entropy bounded unmasking},
  author={Ben-Hamu, Heli and Gat, Itai and Severo, Daniel and Nolte, Niklas S and Karrer, Brian},
  journal={Advances in Neural Information Processing Systems},
  volume={38},
  pages={55981--56007},
  year={2026}
}

@article{austin2021program,
  title={Program synthesis with large language models},
  author={Austin, Jacob and Odena, Augustus and Nye, Maxwell and Bosma, Maarten and Michalewski, Henryk and Dohan, David and Jiang, Ellen and Cai, Carrie and Terry, Michael and Le, Quoc and others},
  journal={arXiv preprint arXiv:2108.07732},
  year={2021}
}

@inproceedings{damani2025learning,
  title={Learning how hard to think: Input-adaptive allocation of lm computation},
  author={Damani, Mehul and Shenfeld, Idan and Peng, Andi and Bobu, Andreea and Andreas, Jacob},
  booktitle={International Conference on Learning Representations},
  volume={2025},
  pages={102783--102802},
  year={2025}
}

@article{chen2021evaluating,
  title={Evaluating large language models trained on code},
  author={Chen, Mark and Tworek, Jerry and Jun, Heewoo and Yuan, Qiming and Pinto, Henrique Ponde De Oliveira and Kaplan, Jared and Edwards, Harri and Burda, Yuri and Joseph, Nicholas and Brockman, Greg and others},
  journal={arXiv preprint arXiv:2107.03374},
  year={2021}
}

@article{chen2026dflash,
  title={Dflash: Block diffusion for flash speculative decoding},
  author={Chen, Jian and Liang, Yesheng and Liu, Zhijian},
  journal={arXiv preprint arXiv:2602.06036},
  year={2026}
}

@article{chang2026dilaserve,
  title={DiLaServe: High SLO Attainment Serving for Diffusion Language Models},
  author={Chang, Tzu-Tao and Hong, Benjamin Yuanyang and Pham, Kiet and Venkataraman, Shivaram},
  journal={arXiv preprint arXiv:2606.29094},
  year={2026}
}

@inproceedings{fan2026dllm,
  title={dLLM-Serve: Bridging the Memory Gap in Diffusion Language Model Serving},
  author={Fan, Jiakun and Zhang, Yanglin and Li, Xiangchen and Nikolopoulos, Dimitrios S},
  booktitle={Proceedings of the 40th ACM International Conference on Supercomputing},
  pages={515--526},
  year={2026}
}

@article{cobbe2021training,
  title={Training verifiers to solve math word problems},
  author={Cobbe, Karl and Kosaraju, Vineet and Bavarian, Mohammad and Chen, Mark and Jun, Heewoo and Kaiser, Lukasz and Plappert, Matthias and Tworek, Jerry and Hilton, Jacob and Nakano, Reiichiro and others},
  journal={arXiv preprint arXiv:2110.14168},
  year={2021}
}

@article{fu2024efficiently,
  title={Efficiently scaling llm reasoning with certaindex},
  author={Fu, Yichao and Chen, Junda and Zhu, Siqi and Fu, Zheyu and Dai, Zhongdongming and Zhuang, Yonghao and Ma, Yian and Qiao, Aurick and Rosing, Tajana and Stoica, Ion and others},
  journal={arXiv preprint arXiv:2412.20993},
  year={2024}
}

@article{zhang2026cost,
  title={Cost-Aware Diffusion Draft Trees for Speculative Decoding},
  author={Zhang, Shuai and Qiu, Huachuan and He, Hongliang and Dai, Yong},
  journal={arXiv preprint arXiv:2606.01813},
  year={2026}
}

@article{hendrycks2021measuring,
  title={Measuring mathematical problem solving with the math dataset},
  author={Hendrycks, Dan and Burns, Collin and Kadavath, Saurav and Arora, Akul and Basart, Steven and Tang, Eric and Song, Dawn and Steinhardt, Jacob},
  journal={arXiv preprint arXiv:2103.03874},
  year={2021}
}

@misc{eval-harness,
  author       = {Gao, Leo and Tow, Jonathan and Abbasi, Baber and Biderman, Stella and Black, Sid and DiPofi, Anthony and Foster, Charles and Golding, Laurence and Hsu, Jeffrey and Le Noac'h, Alain and Li, Haonan and McDonell, Kyle and Muennighoff, Niklas and Ociepa, Chris and Phang, Jason and Reynolds, Laria and Schoelkopf, Hailey and Skowron, Aviya and Sutawika, Lintang and Tang, Eric and Thite, Anish and Wang, Ben and Wang, Kevin and Zou, Andy},
  title        = {The Language Model Evaluation Harness},
  month        = 07,
  year         = 2024,
  publisher    = {Zenodo},
  version      = {v0.4.3},
  doi          = {10.5281/zenodo.12608602},
  url          = {https://zenodo.org/records/12608602}
}

@inproceedings{hong2026improving,
  title={Improving discrete diffusion unmasking policies beyond explicit reference policies},
  author={Hong, Chunsan and An, Seonho and Kim, Min-Soo and Ye, Jong Chul},
  booktitle={International Conference on Learning Representations},
  volume={2026},
  pages={108226--108260},
  year={2026}
}

@article{hu2026elf,
  title={Elf: Embedded language flows},
  author={Hu, Keya and Qiu, Linlu and Lu, Yiyang and Zhao, Hanhong and Li, Tianhong and Kim, Yoon and Andreas, Jacob and He, Kaiming},
  journal={arXiv preprint arXiv:2605.10938},
  year={2026}
}

@article{israel2026accelerating,
  title={Accelerating diffusion llms via adaptive parallel decoding},
  author={Israel, Daniel and Van den Broeck, Guy and Grover, Aditya},
  journal={Advances in neural information processing systems},
  volume={38},
  pages={52870--52888},
  year={2026}
}

@article{jaiswal2025sageserve,
  title={Sageserve: Optimizing llm serving on cloud data centers with forecast aware auto-scaling},
  author={Jaiswal, Shashwat and Jain, Kunal and Simmhan, Yogesh and Parayil, Anjaly and Mallick, Ankur and Wang, Rujia and Amant, Renee St and Bansal, Chetan and Ruhle, Victor and Kulkarni, Anoop and others},
  journal={Proceedings of the ACM on Measurement and Analysis of Computing Systems},
  volume={9},
  number={3},
  pages={1--24},
  year={2025},
  publisher={ACM New York, NY, USA}
}

@article{jazbec2025learning,
  title={Learning unmasking policies for diffusion language models},
  author={Jazbec, Metod and Olausson, Theo X and B{\'e}thune, Louis and Ablin, Pierre and Kirchhof, Michael and Monteiro, Jo{\~a}o and Turrisi, Victor and Ramapuram, Jason and Cuturi, Marco},
  journal={arXiv preprint arXiv:2512.09106},
  year={2025}
}

@article{kim2026klass,
  title={Klass: Kl-guided fast inference in masked diffusion models},
  author={Kim, Seo Hyun and Hong, Sunwoo and Jung, Hojung and Park, Youngrok and Yun, Se-Young},
  journal={Advances in Neural Information Processing Systems},
  volume={38},
  pages={92267--92301},
  year={2026}
}

@inproceedings{li2026diffusion,
  title={Diffusion language model knows the answer before it decodes},
  author={Li, Pengxiang and Zhou, Yefan and Muhtar, Dilxat and Yin, Lu and Yan, Shilin and Shen, Li and Liang, Yi and Vosoughi, Soroush and Liu, Shiwei},
  booktitle={The Fourteenth International Conference on Learning Representations},
  year={2026}
}

@inproceedings{lightman2024let,
  title={Let's verify step by step},
  author={Lightman, Hunter and Kosaraju, Vineet and Burda, Yuri and Edwards, Harrison and Baker, Bowen and Lee, Teddy and Leike, Jan and Schulman, John and Sutskever, Ilya and Cobbe, Karl},
  booktitle={International Conference on Learning Representations},
  volume={2024},
  pages={39578--39601},
  year={2024}
}

@article{nie2026large,
  title={Large language diffusion models},
  author={Nie, Shen and Zhu, Fengqi and You, Zebin and Zhang, Xiaolu and Ou, Jingyang and Hu, Jun and Zhou, Jun and Lin, Yankai and Wen, Ji-Rong and Li, Chongxuan},
  journal={Advances in Neural Information Processing Systems},
  volume={38},
  pages={50608--50646},
  year={2026}
}

@article{ye2508dream,
  title={Dream 7b: Diffusion large language models, 2025},
  author={Ye, Jiacheng and Xie, Zhihui and Zheng, Lin and Gao, Jiahui and Wu, Zirui and Jiang, Xin and Li, Zhenguo and Kong, Lingpeng},
  journal={URL https://arxiv. org/abs/2508.15487},
  year={2025}
}

@article{bie2026llada2,
  title={Llada2. 1: Speeding up text diffusion via token editing},
  author={Bie, Tiwei and Cao, Maosong and Cao, Xiang and Chen, Bingsen and Chen, Fuyuan and Chen, Kun and Du, Lun and Feng, Daozhuo and Feng, Haibo and Gong, Mingliang and others},
  journal={arXiv preprint arXiv:2602.08676},
  year={2026}
}

@misc{openaiscale,
  author       = {{OpenAI}},
  title        = {{Scale Tier} for API Customers},
  howpublished = {OpenAI API documentation},
  year         = {2026},
  note         = {Accessed October 9, 2026},
  url          = {https://openai.com/api-scale-tier/}
}

@inproceedings{bao2026learning,
  title={Learning to parallel: Accelerating diffusion large language models via learnable parallel decoding},
  author={Bao, Wenrui and Chen, Zhiben and Xu, Dan and Shang, Yuzhang},
  booktitle={International Conference on Learning Representations},
  volume={2026},
  pages={83376--83392},
  year={2026}
}

@inproceedings{wang2026creditdecoding,
  title={CreditDecoding: Accelerating Parallel Decoding in Diffusion Large Language Models with Trace Credit},
  author={Wang, Kangyu and Jiang, Zhiyun and Feng, Haibo and Zhao, Weijia and Liu, Lin and Li, Jianguo and Lan, Zhenzhong and Lin, Weiyao},
  booktitle={Proceedings of the 64th Annual Meeting of the Association for Computational Linguistics (Volume 1: Long Papers)},
  pages={11105--11123},
  year={2026}
}

@inproceedings{wei2026accelerating,
  title={Accelerating diffusion large language models with slowfast sampling: The three golden principles},
  author={Wei, Qingyan and Zhang, Yaojie and Liu, Zhiyuan and Zeng, Puyu and Wang, Yuxuan and Qi, Biqing and Liu, Dongrui and Zhang, Linfeng},
  booktitle={International Conference on Learning Representations},
  volume={2026},
  pages={4340--4354},
  year={2026}
}

@article{schuster2022confident,
  title={Confident adaptive language modeling},
  author={Schuster, Tal and Fisch, Adam and Gupta, Jai and Dehghani, Mostafa and Bahri, Dara and Tran, Vinh and Tay, Yi and Metzler, Donald},
  journal={Advances in Neural Information Processing Systems},
  volume={35},
  pages={17456--17472},
  year={2022}
}

@inproceedings{snell2025scaling,
  title={Scaling LLM test-time compute optimally can be more effective than scaling parameters for reasoning},
  author={Snell, Charlie and Lee, Jaehoon and Xu, Kelvin and Kumar, Aviral},
  booktitle={International Conference on Learning Representations},
  volume={2025},
  pages={10131--10165},
  year={2025}
}

@inproceedings{zuo2026strategic,
  title={Strategic scaling of test-time compute: A bandit learning approach},
  author={Zuo, Bowen and Zhu, Yinglun},
  booktitle={International Conference on Learning Representations},
  volume={2026},
  pages={134105--134128},
  year={2026}
}

@article{cheng2026dspark,
  title={DSpark: Confidence-Scheduled Speculative Decoding with Semi-Autoregressive Generation},
  author={Cheng, Xin and Yu, Xingkai and Shao, Chenze and Li, Jiashi and Xiong, Yunfan and Qian, Yi and Zhu, Jiaqi and Ma, Shirong and Zhang, Xiaokang and Ye, Jiasheng and others},
  journal={arXiv preprint arXiv:2607.05147},
  year={2026}
}

@inproceedings{wu2026dynamic,
  title={Dynamic-dLLM: Dynamic cache-budget and adaptive parallel decoding for training-free acceleration of diffusion LLM},
  author={Wu, Tianyi and Sun, Xiaoxi and Jiao, Yanhua and Li, Yulin and Chen, Yixin and Cao, Yun-Hao and Hu, Yi-Qi and Tian, Zhuotao},
  booktitle={International Conference on Learning Representations},
  volume={2026},
  pages={66347--66365},
  year={2026}
}

@inproceedings{wu2026fast,
  title={Fast-dllm: Training-free acceleration of diffusion llm by enabling kv cache and parallel decoding},
  author={Wu, Chengyue and Zhang, Hao and Xue, Shuchen and Liu, Zhijian and Diao, Shizhe and Zhu, Ligeng and Luo, Ping and Han, Song and Xie, Enze},
  booktitle={International Conference on Learning Representations},
  volume={2026},
  pages={57027--57051},
  year={2026}
}

@inproceedings{yu2022orca,
  title={Orca: A distributed serving system for $\{$Transformer-Based$\}$ generative models},
  author={Yu, Gyeong-In and Jeong, Joo Seong and Kim, Geon-Woo and Kim, Soojeong and Chun, Byung-Gon},
  booktitle={16th USENIX symposium on operating systems design and implementation (OSDI 22)},
  pages={521--538},
  year={2022}
}

@inproceedings{agrawal2024taming,
  title={Taming $\{$Throughput-Latency$\}$ tradeoff in $\{$LLM$\}$ inference with $\{$Sarathi-Serve$\}$},
  author={Agrawal, Amey and Kedia, Nitin and Panwar, Ashish and Mohan, Jayashree and Kwatra, Nipun and Gulavani, Bhargav and Tumanov, Alexey and Ramjee, Ramachandran},
  booktitle={18th USENIX symposium on operating systems design and implementation (OSDI 24)},
  pages={117--134},
  year={2024}
}

@article{zhou2023instruction,
  title={Instruction-following evaluation for large language models},
  author={Zhou, Jeffrey and Lu, Tianjian and Mishra, Swaroop and Brahma, Siddhartha and Basu, Sujoy and Luan, Yi and Zhou, Denny and Hou, Le},
  journal={arXiv preprint arXiv:2311.07911},
  year={2023}
}

@inproceedings{zhu2026llada,
  title={Llada 1.5: Variance-reduced preference optimization for large language diffusion models},
  author={Zhu, Fengqi and Wang, Rongzhen and Nie, Shen and Zhang, Xiaolu and Wu, Chunwei and Zhou, Jun and Lin, Yankai and Wen, Ji-Rong and Li, Chongxuan},
  booktitle={Proceedings of the 64th Annual Meeting of the Association for Computational Linguistics (Volume 1: Long Papers)},
  pages={11425--11460},
  year={2026}
}

@article{zhang2026low,
  title={Low Perplexity is Repetition: A One-Dimensional Self-Conditioning Attractor in Continuous Diffusion LMs},
  author={Zhang, Shuai and Chen, Zijie and He, Hongliang and Du, Lun and Lan, Zhenzhong},
  journal={arXiv preprint arXiv:2607.00588},
  year={2026}
}

\clearpage
\appendix
\section{Additional Experimental Results}
\label{app:supplementary-results}

\subsection{Request-Level Analysis on LLaDA-8B-Instruct}
\label{app:allocation-dynamics-8b}

Figure~\ref{fig:allocation-evidence-8b} extends the request-level analysis in
Section~\ref{sec:allocation-dynamics} to LLaDA-8B-Instruct on the same benchmark mixture. It shows
the same qualitative pattern: Batch-Paced changes commit allocation across requests and denoising
steps, with the clearest confidence shifts and group-level performance gains among lower-confidence
requests.

\subsection{Robustness Analyses}
\label{app:robustness-analyses}

\subsubsection{Task-Mixture Ratios}
\label{sec:ratio-sensitivity}

To test sensitivity to two-task batch composition, we vary the GSM8K:HumanEval ratio while holding
the decoding configuration and target budgets fixed. For each ratio and seed, we form 128 examples
and partition them into eight batches of 16. Integer task counts are assigned per batch using the
largest-remainder rule, and examples are sampled without replacement from the benchmark pools.
Each decoding policy uses the same examples in the same batch order; seeds vary the example
membership and batch arrangement. Table~\ref{tab:ratio} shows that Batch-Paced has higher mean
performance across the tested ratios, although the gains do not increase monotonically with the ratio.

\begin{table}[H]
\centering
\footnotesize
\setlength{\tabcolsep}{3pt}
\renewcommand{\arraystretch}{1.08}
\caption{\textbf{Batch-Paced improves performance across task-mixture ratios.} The table compares both policies across five GSM8K--HumanEval ratios, backbones, and target budgets.}
\label{tab:ratio}
\newcommand{\ratioacc}[2]{#1_{\scriptscriptstyle\,#2}}
\begin{tabular*}{\columnwidth}{@{}>{\centering\arraybackslash}p{0.48in}@{\extracolsep{\fill}}cccc@{}}
\toprule
\multirow{3}{*}{\raisebox{-11pt}{\makebox[0.48in][c]{Ratio}}} & \multicolumn{2}{c}{Avg. budget=64} & \multicolumn{2}{c}{Avg. budget=128} \\
\cmidrule(lr){2-3} \cmidrule(lr){4-5}
 & Uniform & Batch- & Uniform & Batch- \\
 & baseline & Paced & baseline & Paced \\
\midrule
\multicolumn{5}{l}{\hspace{3pt}LLaDA-8B-Instruct} \\
\cmidrule(lr){1-5}
1:7 & $\ratioacc{28.3}{1.4}$ & {\boldmath$\ratioacc{40.9}{0.9}^{*}$} & $\ratioacc{34.2}{0.9}$ & {\boldmath$\ratioacc{48.4}{2.1}^{*}$} \\
1:3 & $\ratioacc{34.4}{1.0}$ & {\boldmath$\ratioacc{48.8}{2.7}^{*}$} & $\ratioacc{43.0}{0.8}$ & {\boldmath$\ratioacc{56.7}{1.6}^{*}$} \\
1:1 & $\ratioacc{46.1}{0.0}$ & {\boldmath$\ratioacc{58.4}{1.7}^{*}$} & $\ratioacc{55.5}{0.0}$ & {\boldmath$\ratioacc{66.7}{0.9}^{*}$} \\
3:1 & $\ratioacc{55.8}{2.4}$ & {\boldmath$\ratioacc{71.4}{4.7}^{*}$} & $\ratioacc{67.7}{2.0}$ & {\boldmath$\ratioacc{74.7}{2.0}^{*}$} \\
7:1 & $\ratioacc{62.5}{1.0}$ & {\boldmath$\ratioacc{78.1}{4.0}^{*}$} & $\ratioacc{73.1}{2.0}$ & {\boldmath$\ratioacc{80.8}{2.1}^{*}$} \\
\midrule
\multicolumn{5}{l}{\hspace{3pt}LLaDA 1.5} \\
\cmidrule(lr){1-5}
1:7 & $\ratioacc{31.4}{1.3}$ & {\boldmath$\ratioacc{46.7}{3.4}^{*}$} & $\ratioacc{40.2}{1.3}$ & {\boldmath$\ratioacc{53.4}{1.0}^{*}$} \\
1:3 & $\ratioacc{37.0}{1.2}$ & {\boldmath$\ratioacc{53.8}{1.0}^{*}$} & $\ratioacc{49.5}{0.9}$ & {\boldmath$\ratioacc{59.8}{1.6}^{*}$} \\
1:1 & $\ratioacc{46.9}{0.0}$ & {\boldmath$\ratioacc{62.8}{3.4}^{*}$} & $\ratioacc{61.7}{0.0}$ & {\boldmath$\ratioacc{69.5}{1.0}^{*}$} \\
3:1 & $\ratioacc{58.3}{1.6}$ & {\boldmath$\ratioacc{73.4}{5.4}^{*}$} & $\ratioacc{70.2}{1.5}$ & {\boldmath$\ratioacc{76.6}{2.6}^{*}$} \\
7:1 & $\ratioacc{60.5}{1.3}$ & {\boldmath$\ratioacc{78.1}{4.0}^{*}$} & $\ratioacc{73.1}{2.0}$ & {\boldmath$\ratioacc{80.8}{2.1}^{*}$} \\
\bottomrule
\end{tabular*}
\end{table}

\subsubsection{Budget and Decoding-Configuration Trends}
\label{app:decoding-choices}

\begin{figure}[t]
\centering
\includegraphics[width=\columnwidth]{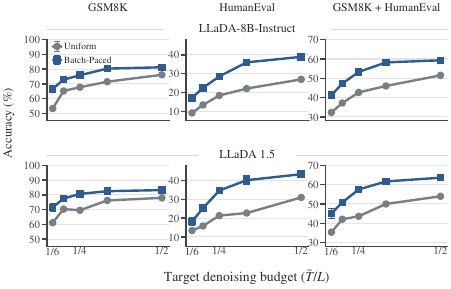}
\caption{\textbf{Batch-Paced outperforms Uniform at every tested average budget.} Accuracy as the
average denoising budget varies across task settings and backbones.}
\label{fig:budget-sensitivity}
\end{figure}

To test a wider range of average denoising budgets, we use $B=16$ and $L=256$ and vary the integer
average commit rate $r\in\{2,3,4,5,6\}$. The corresponding target budget fraction is $\bar T/L=1/r$;
for $L=256$, $r=4$ and $r=2$ correspond to average budgets of 64 and 128, respectively. When the
block length is not divisible by $r$, the final pass commits fewer than $r$ remaining positions.
Figure~\ref{fig:budget-sensitivity} reports accuracy on GSM8K, HumanEval, and their mixture separately
for each model. Batch-Paced has higher mean accuracy than Uniform at every tested rate in all six
model--task settings, while the absolute accuracy trajectories vary by setting.

\begin{figure*}[t]
\centering
\includegraphics[width=0.99\textwidth]{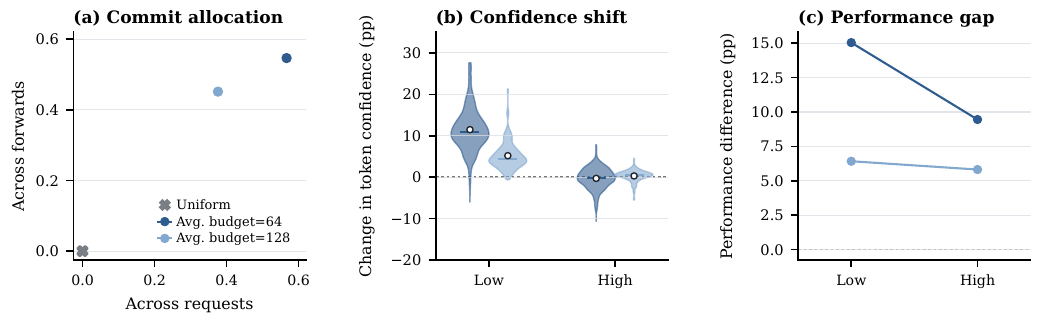}
\caption{\textbf{LLaDA-8B-Instruct shows the same pattern, with its largest confidence shifts and performance gains in the low-confidence group.}
The panels show (a) variation in commit allocation across requests and forwards, (b) confidence shifts by
Uniform-confidence group, and (c) paired performance differences between Batch-Paced and Uniform.}
\label{fig:allocation-evidence-8b}
\vspace{0.7em}
\begin{minipage}[t]{0.49\textwidth}
\vspace{0pt}\centering
\includegraphics[width=\linewidth]{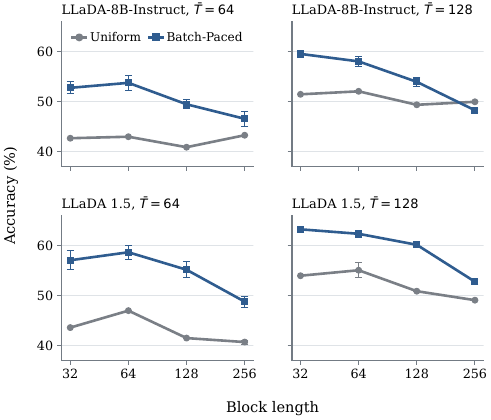}
\captionof{figure}{\textbf{Batch-Paced's advantage largely persists across block lengths.} The curves compare
accuracy under Uniform and Batch-Paced as block length varies.}
\label{fig:block-sensitivity}
\end{minipage}\hfill
\begin{minipage}[t]{0.49\textwidth}
\vspace{0pt}\centering
\includegraphics[width=\linewidth]{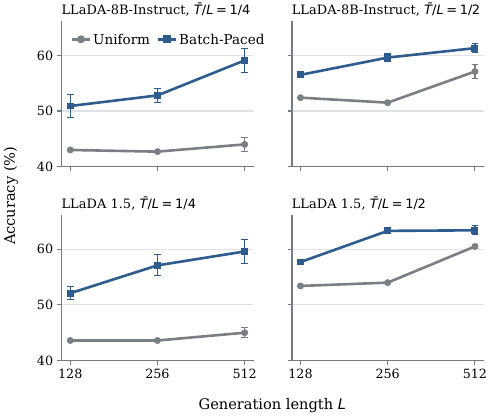}
\captionof{figure}{\textbf{Batch-Paced maintains higher performance across generation lengths.} Curves compare
Uniform and Batch-Paced results for both LLaDA models.}
\label{fig:genlength-sensitivity}
\end{minipage}
\end{figure*}

Figures~\ref{fig:block-sensitivity} and~\ref{fig:genlength-sensitivity} give the full block-length
and generation-length trends underlying the representative comparisons in
Table~\ref{tab:sensitivity-summary}.

\subsection{Fast-dLLM Budget-Enforcement Variants}
\label{sec:allocation-variants}

Section~\ref{sec:fastdllm} introduces two stopping-rule variants that retain Fast-dLLM's
sequence-wise threshold selection. The separate-sequence variant gives each sequence its own
allowance of $\bar T$ denoising forward passes and force-completes its remaining masks when that
allowance is exhausted. The shared-batch variant draws from a pool of $B\bar T$ units, charging each
forward pass by the number of unfinished sequences; when the pool cannot fund another pass, it
force-completes all unfinished sequences. Thus, only the latter transfers unused allowance across
sequences, while neither replaces Fast-dLLM's threshold-based token selection.

For each variant, model, benchmark, budget, and seed, we repeat the binary search described in
Section~\ref{sec:fastdllm}, using the same sampled examples and batch order as the main comparison.
Among tested thresholds, we select the feasible one whose largest per-batch mean denoising budget is
closest to the target; forced-completion forwards count toward this budget. Table~\ref{tab:allocation-variants}
reports the results. Batch-Paced has higher mean performance than either variant across the displayed
benchmark mixtures, while comparisons with unmodified Fast-dLLM remain mixed. This indicates that
adding budget enforcement or sharing to threshold decoding does not by itself reproduce Batch-Paced's
mixture results.

\begin{table*}[t]
\centering
\footnotesize
\setlength{\tabcolsep}{1pt}
\renewcommand{\arraystretch}{0.98}
\caption{\textbf{Batch-Paced usually outperforms the budget-enforcement variants on benchmark mixtures.} The table compares Fast-dLLM, its two runtime budget-enforcement variants, and Batch-Paced on individual benchmarks and mixtures across both backbones and two average denoising budgets.}
\label{tab:allocation-variants}
\begin{tabular}{@{}>{\raggedright\arraybackslash}p{0.95in}>{\raggedright\arraybackslash}p{2.15in}*{4}{>{\centering\arraybackslash}p{0.75in}}@{}}
\toprule
\multirow{2}{*}{Benchmark} & \multirow{2}{*}{Method} & \multicolumn{2}{c}{LLaDA-8B-Instruct} & \multicolumn{2}{c}{LLaDA 1.5} \\
\cmidrule(lr){3-4} \cmidrule(lr){5-6}
 & & $\bar T=64$ & $\bar T=128$ & $\bar T=64$ & $\bar T=128$ \\
\midrule
\multirow{4}{*}{HumanEval} & Fast-dLLM & \ensuremath{36.6_{\scriptscriptstyle\,1.9}} & {\boldmath\ensuremath{40.9_{\scriptscriptstyle\,0.0}}} & \ensuremath{33.0_{\scriptscriptstyle\,11.5}} & {\boldmath\ensuremath{45.7_{\scriptscriptstyle\,0.0}}} \\
 & Fast-dLLM + separate sequence budgets & {\boldmath\ensuremath{39.9_{\scriptscriptstyle\,1.5}}} & \ensuremath{40.2_{\scriptscriptstyle\,0.0}} & {\boldmath\ensuremath{41.3_{\scriptscriptstyle\,2.1}}} & \ensuremath{45.1_{\scriptscriptstyle\,0.0}} \\
 & Fast-dLLM + shared batch budget & \ensuremath{19.9_{\scriptscriptstyle\,10.7}} & {\boldmath\ensuremath{40.9_{\scriptscriptstyle\,0.0}}} & \ensuremath{32.7_{\scriptscriptstyle\,11.2}} & {\boldmath\ensuremath{45.7_{\scriptscriptstyle\,0.0}}} \\
 & \textbf{Batch-Paced (ours)} & \ensuremath{29.3_{\scriptscriptstyle\,1.2}} & \ensuremath{38.4_{\scriptscriptstyle\,0.4}} & \ensuremath{34.6_{\scriptscriptstyle\,0.9}} & \ensuremath{43.2_{\scriptscriptstyle\,0.9}} \\
\midrule
\multirow{4}{*}{IFEval} & Fast-dLLM & \ensuremath{39.6_{\scriptscriptstyle\,1.9}} & \ensuremath{50.5_{\scriptscriptstyle\,1.9}} & \ensuremath{40.4_{\scriptscriptstyle\,1.6}} & \ensuremath{51.7_{\scriptscriptstyle\,0.7}} \\
 & Fast-dLLM + separate sequence budgets & \ensuremath{35.6_{\scriptscriptstyle\,9.1}} & \ensuremath{43.7_{\scriptscriptstyle\,3.0}} & \ensuremath{38.6_{\scriptscriptstyle\,9.3}} & \ensuremath{48.3_{\scriptscriptstyle\,1.6}} \\
 & Fast-dLLM + shared batch budget & \ensuremath{27.3_{\scriptscriptstyle\,10.9}} & \ensuremath{44.4_{\scriptscriptstyle\,14.1}} & \ensuremath{29.3_{\scriptscriptstyle\,10.2}} & \ensuremath{27.9_{\scriptscriptstyle\,13.7}} \\
 & \textbf{Batch-Paced (ours)} & {\boldmath\ensuremath{43.8_{\scriptscriptstyle\,1.7}}} & {\boldmath\ensuremath{51.7_{\scriptscriptstyle\,0.8}}} & {\boldmath\ensuremath{47.7_{\scriptscriptstyle\,0.8}}} & {\boldmath\ensuremath{54.5_{\scriptscriptstyle\,0.7}}} \\
\midrule
\multirow{4}{*}{GSM8K} & Fast-dLLM & {\boldmath\ensuremath{78.8_{\scriptscriptstyle\,0.4}}} & {\boldmath\ensuremath{81.2_{\scriptscriptstyle\,0.0}}} & {\boldmath\ensuremath{81.6_{\scriptscriptstyle\,0.9}}} & \ensuremath{83.3_{\scriptscriptstyle\,0.1}} \\
 & Fast-dLLM + separate sequence budgets & \ensuremath{64.4_{\scriptscriptstyle\,6.1}} & \ensuremath{72.3_{\scriptscriptstyle\,0.1}} & \ensuremath{71.1_{\scriptscriptstyle\,5.3}} & \ensuremath{76.1_{\scriptscriptstyle\,0.2}} \\
 & Fast-dLLM + shared batch budget & \ensuremath{77.2_{\scriptscriptstyle\,5.0}} & {\boldmath\ensuremath{81.2_{\scriptscriptstyle\,0.0}}} & \ensuremath{80.7_{\scriptscriptstyle\,0.0}} & \ensuremath{83.3_{\scriptscriptstyle\,0.1}} \\
 & \textbf{Batch-Paced (ours)} & \ensuremath{77.5_{\scriptscriptstyle\,1.1}} & \ensuremath{81.0_{\scriptscriptstyle\,0.5}} & \ensuremath{80.5_{\scriptscriptstyle\,0.8}} & {\boldmath\ensuremath{83.8_{\scriptscriptstyle\,0.6}}} \\
\midrule
\multirow{4}{*}{\shortstack[l]{GSM8K +\\IFEval}} & Fast-dLLM & \ensuremath{46.0_{\scriptscriptstyle\,2.7}} & \ensuremath{68.8_{\scriptscriptstyle\,0.7}} & \ensuremath{45.0_{\scriptscriptstyle\,3.7}} & \ensuremath{71.2_{\scriptscriptstyle\,0.9}} \\
 & Fast-dLLM + separate sequence budgets & \ensuremath{55.2_{\scriptscriptstyle\,1.8}} & \ensuremath{53.6_{\scriptscriptstyle\,5.8}} & \ensuremath{59.5_{\scriptscriptstyle\,1.2}} & \ensuremath{59.0_{\scriptscriptstyle\,4.7}} \\
 & Fast-dLLM + shared batch budget & \ensuremath{39.1_{\scriptscriptstyle\,5.9}} & {\boldmath\ensuremath{69.1_{\scriptscriptstyle\,0.2}}} & \ensuremath{39.1_{\scriptscriptstyle\,0.6}} & \ensuremath{68.2_{\scriptscriptstyle\,4.2}} \\
 & \textbf{Batch-Paced (ours)} & {\boldmath\ensuremath{58.2_{\scriptscriptstyle\,1.2}}} & \ensuremath{68.3_{\scriptscriptstyle\,1.0}} & {\boldmath\ensuremath{62.9_{\scriptscriptstyle\,0.8}}} & {\boldmath\ensuremath{71.7_{\scriptscriptstyle\,0.4}}} \\
\midrule
\multirow{4}{*}{\shortstack[l]{GSM8K +\\HumanEval}} & Fast-dLLM & {\boldmath\ensuremath{57.9_{\scriptscriptstyle\,0.1}}} & {\boldmath\ensuremath{60.7_{\scriptscriptstyle\,0.0}}} & {\boldmath\ensuremath{60.6_{\scriptscriptstyle\,1.4}}} & \ensuremath{64.3_{\scriptscriptstyle\,0.1}} \\
 & Fast-dLLM + separate sequence budgets & \ensuremath{52.3_{\scriptscriptstyle\,2.6}} & \ensuremath{55.9_{\scriptscriptstyle\,1.5}} & \ensuremath{54.9_{\scriptscriptstyle\,4.0}} & \ensuremath{60.8_{\scriptscriptstyle\,0.2}} \\
 & Fast-dLLM + shared batch budget & \ensuremath{44.8_{\scriptscriptstyle\,7.3}} & {\boldmath\ensuremath{60.7_{\scriptscriptstyle\,0.0}}} & \ensuremath{53.5_{\scriptscriptstyle\,7.4}} & {\boldmath\ensuremath{64.3_{\scriptscriptstyle\,0.2}}} \\
 & \textbf{Batch-Paced (ours)} & \ensuremath{52.8_{\scriptscriptstyle\,1.2}} & \ensuremath{59.6_{\scriptscriptstyle\,0.7}} & \ensuremath{57.1_{\scriptscriptstyle\,1.9}} & \ensuremath{63.3_{\scriptscriptstyle\,0.5}} \\
\midrule
\multirow{4}{*}{\shortstack[l]{HumanEval +\\IFEval}} & Fast-dLLM & \ensuremath{24.0_{\scriptscriptstyle\,2.8}} & \ensuremath{46.2_{\scriptscriptstyle\,3.0}} & \ensuremath{27.2_{\scriptscriptstyle\,4.3}} & \ensuremath{48.5_{\scriptscriptstyle\,5.7}} \\
 & Fast-dLLM + separate sequence budgets & \ensuremath{30.9_{\scriptscriptstyle\,6.6}} & \ensuremath{41.3_{\scriptscriptstyle\,0.2}} & \ensuremath{35.4_{\scriptscriptstyle\,5.5}} & \ensuremath{45.9_{\scriptscriptstyle\,0.2}} \\
 & Fast-dLLM + shared batch budget & \ensuremath{18.8_{\scriptscriptstyle\,6.1}} & \ensuremath{40.9_{\scriptscriptstyle\,8.5}} & \ensuremath{18.9_{\scriptscriptstyle\,5.5}} & \ensuremath{45.1_{\scriptscriptstyle\,7.7}} \\
 & \textbf{Batch-Paced (ours)} & {\boldmath\ensuremath{32.7_{\scriptscriptstyle\,2.0}}} & {\boldmath\ensuremath{46.8_{\scriptscriptstyle\,1.3}}} & {\boldmath\ensuremath{36.3_{\scriptscriptstyle\,1.8}}} & {\boldmath\ensuremath{50.1_{\scriptscriptstyle\,1.3}}} \\
\midrule
\multirow{4}{*}{\shortstack[l]{GSM8K +\\HumanEval +\\IFEval}} & Fast-dLLM & \ensuremath{33.9_{\scriptscriptstyle\,5.6}} & \ensuremath{59.3_{\scriptscriptstyle\,1.0}} & \ensuremath{36.6_{\scriptscriptstyle\,2.3}} & \ensuremath{61.8_{\scriptscriptstyle\,1.2}} \\
 & Fast-dLLM + separate sequence budgets & \ensuremath{44.1_{\scriptscriptstyle\,5.1}} & \ensuremath{50.9_{\scriptscriptstyle\,0.1}} & \ensuremath{50.0_{\scriptscriptstyle\,2.8}} & \ensuremath{55.9_{\scriptscriptstyle\,0.1}} \\
 & Fast-dLLM + shared batch budget & \ensuremath{24.6_{\scriptscriptstyle\,12.4}} & {\boldmath\ensuremath{59.3_{\scriptscriptstyle\,1.0}}} & \ensuremath{25.3_{\scriptscriptstyle\,9.7}} & \ensuremath{61.6_{\scriptscriptstyle\,0.5}} \\
 & \textbf{Batch-Paced (ours)} & {\boldmath\ensuremath{48.3_{\scriptscriptstyle\,1.1}}} & \ensuremath{57.3_{\scriptscriptstyle\,0.9}} & {\boldmath\ensuremath{51.2_{\scriptscriptstyle\,2.4}}} & {\boldmath\ensuremath{61.9_{\scriptscriptstyle\,0.8}}} \\
\bottomrule
\end{tabular}
\end{table*}

\subsection{Depth Trade-Off on LLaDA-8B-Instruct}
\label{app:round-allowance-8b}

\begin{figure*}[t]
\centering
\includegraphics[width=0.99\textwidth]{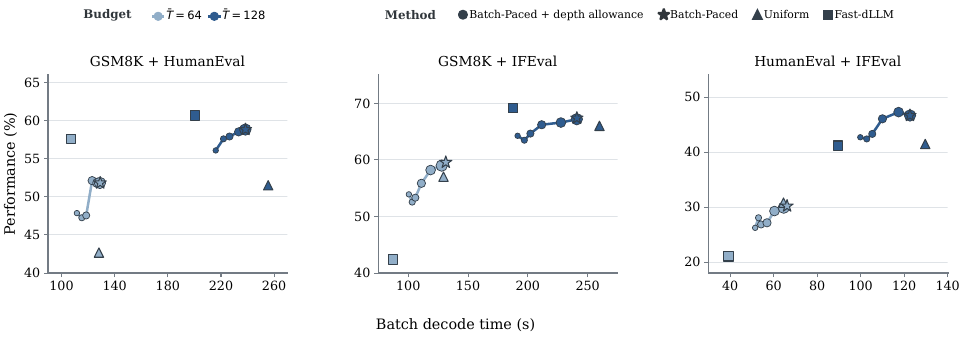}
\caption{\textbf{Depth limits trade Batch-Paced performance for shorter decode time on
LLaDA-8B-Instruct.} The plot compares Uniform, Fast-dLLM, and original and depth-limited Batch-Paced
on the same three task mixtures and target budgets as Figure~\ref{fig:round-allowance}; larger circles
denote larger depth allowances.}
\label{fig:round-allowance-8b}
\end{figure*}

Figure~\ref{fig:round-allowance-8b} gives the corresponding time--performance results for
LLaDA-8B-Instruct. The same trade-off is visible: tighter depth allowances shorten batch decoding,
while larger allowances move performance and decode time toward the original policy.

\section{Decoding-Depth-Allowance Algorithm}
\label{sec:round-capped}

Section~\ref{sec:depth-tradeoff} introduces a decoding-depth allowance that bounds the batch to
$D_{\max}$ sequential denoising steps while retaining Batch-Paced's shared budget and batch-wide
confidence ranking. Each step is one batched denoising forward pass; wall-clock decode time is
measured separately. At a decision point, let $\mathcal E_i$ contain the masked positions $(i,j)$
in the current block of sequence $i$, let $\mathcal E=\bigcup_{i\in\mathcal A}\mathcal E_i$, and
let $m_i=|\mathcal E_i|$. Let $n_i$ count the denoising forward passes used by sequence $i$ before
the current scoring pass. Let
$N_{\mathrm{blk}}=\lceil L/H\rceil$ be the number of generation blocks, and let $g_i$ count the
blocks after sequence $i$'s current block; each requires at least one forward pass.  Thus,
$g_i=N_{\mathrm{blk}}-1$ initially and decreases by one whenever sequence $i$ advances to its next
block.  The current block therefore has
\begin{compactequation}
 q_i = D_{\max}-n_i-g_i
 \label{eq:round-cap-slack}
\end{compactequation}
forward opportunities remaining, including the current pass whose confidence scores are being
processed. For every unfinished sequence, the allowance must leave at least one forward opportunity,
so $q_i\geq 1$. To finish the current block within its remaining forward opportunities, sequence
$i$ receives a minimum commit floor
\begin{compactequation}
 f_i = \max\!\left(1,\left\lceil\frac{m_i}{q_i}\right\rceil\right),
 \label{eq:round-cap-floor}
\end{compactequation}
provided $q_i\geq 1$. The floor rises as a sequence has fewer denoising steps left to commit its
masked positions; the required positions are its highest-confidence masked positions.

Let $K$ be the commit count selected by the shared-budget rule in
Eq.~\ref{eq:batch-paced-k}.  The policy first selects the pooled top-$K$ candidates:
\begin{compactequation}
 \mathcal C_0 = \operatorname{Top}_{K}(\mathcal E; s_{ij}),
 \label{eq:round-cap-selection}
\end{compactequation}
then adds, for each sequence $i$, the highest-confidence positions in $\mathcal E_i$ needed to
raise $|\mathcal C_0\cap\mathcal E_i|$ to $f_i$.  Thus, each unfinished sequence commits at
least one token, while all commits beyond these per-sequence floors follow the global ranking.
Progressively rising floors schedule the commits needed to meet the depth allowance across the
remaining denoising steps. Every step counts toward both the
decoding-depth allowance and the shared compute pool. The algorithm assumes a feasible setting:
$R\geq|\mathcal A|$ and $q_i\geq1$ at each step. Algorithm~\ref{alg:round-capped} gives the full
loop; without a decoding-depth allowance, its selector reduces to the base policy in
Algorithm~\ref{alg:batch-paced}.

\begin{algorithm*}[t]
\caption{Batch-Paced decoding with a decoding-depth allowance}
\label{alg:round-capped}
\begin{algorithmic}[1]
\Require batch $\mathcal B$, generation length $L$, block length $H$, average budget $\bar T$,
         optional feasible decoding-depth allowance $D_{\max}$ (in denoising steps)
\State $N_{\mathrm{blk}}\gets\lceil L/H\rceil$
\State $R\gets|\mathcal B|\bar T$
\State Initialize masked generations
\State Set $n_i\gets0$ and $g_i\gets N_{\mathrm{blk}}-1$ for all $i$
\While{some sequence remains unfinished}
  \State $\mathcal A\gets$ unfinished sequences
  \State Form current candidate sets $\mathcal E_i$ for $i\in\mathcal A$
  \State $m_i\gets|\mathcal E_i|$ for all $i\in\mathcal A$
  \State $\mathcal E\gets\bigcup_{i\in\mathcal A}\mathcal E_i$
  \State Set $f_i\gets1$ for all $i\in\mathcal A$
  \If{decoding-depth allowance is imposed}
    \ForAll{$i\in\mathcal A$}
      \State $q_i\gets D_{\max}-n_i-g_i$
      \State $f_i\gets\max(1,\lceil m_i/q_i\rceil)$
    \EndFor
  \EndIf
  \State Run one batched denoising forward pass
  \State Compute scores $s_{ij}$ over $\mathcal E$
  \State Compute $K$ using Eq.~\ref{eq:batch-paced-k}
  \State $\mathcal C\gets\operatorname{Top}_{K}(\mathcal E;s_{ij})$
  \ForAll{$i\in\mathcal A$}
    \While{$|\mathcal C\cap\mathcal E_i|<f_i$}
      \State Add $\arg\max_{(i,j)\in\mathcal E_i\setminus\mathcal C}s_{ij}$ to $\mathcal C$
    \EndWhile
  \EndFor
  \State Commit $\mathcal C$
  \State $R\gets R-|\mathcal A|$; increment $n_i$ for all $i\in\mathcal A$
  \State Advance completed blocks; decrement $g_i$ for sequences entering a new block
\EndWhile
\State \Return completed batch
\end{algorithmic}
\end{algorithm*}

\end{document}